\pdfoutput=1

\documentclass[11pt]{article}

\usepackage[final]{acl}

\usepackage{times}
\usepackage{latexsym}
\usepackage{arydshln}
\usepackage[T1]{fontenc}
\usepackage[utf8]{inputenc}
\usepackage{microtype}
\usepackage{inconsolata}
\usepackage{graphicx}
\usepackage{booktabs}
\usepackage{multirow}
\usepackage{array} 
\usepackage{subcaption}
\usepackage[table]{xcolor}

\title{On Multilingual Encoder Language Model Compression \\ for Low-Resource Languages}

\author{Daniil Gurgurov\textsuperscript{\normalfont1,2} \quad Michal Gregor\textsuperscript{\normalfont3} \quad Josef van Genabith\textsuperscript{\normalfont1,2} \quad Simon Ostermann\textsuperscript{\normalfont1,2,4}\\
    \textsuperscript{1}Saarland University\\ 
    \textsuperscript{2}German Research Center for Artificial Intelligence (DFKI)\\
    \textsuperscript{3}Kempelen Institute of Intelligent Technologies (KInIT)\\
    \textsuperscript{4}Centre for European Research in Trusted AI (CERTAIN)\\
    {\small \texttt{ \{daniil.gurgurov, josef.van\_genabith, simon.ostermann\}@dfki.de, michal.gregor@kinit.sk}}
    }

\begin{document}
\maketitle

\begin{abstract}
In this paper, we combine two-step knowledge distillation, structured pruning, truncation, and vocabulary trimming for extremely compressing multilingual encoder-only language models for low-resource languages. Our novel approach systematically combines existing techniques and takes them to the extreme, reducing layer depth, feed-forward hidden size, and intermediate layer embedding size to create significantly smaller monolingual models while retaining essential language-specific knowledge. \textbf{We achieve compression rates of up to 92\% while maintaining competitive performance, with average drops of 2–10\% for moderate compression and 8–13\% at maximum compression} in four downstream tasks, including sentiment analysis, topic classification, named entity recognition, and part-of-speech tagging, across three low-resource languages. Notably, the performance degradation correlates with the amount of language-specific data in the teacher model, with larger datasets resulting in smaller performance losses. Additionally, we conduct ablation studies to identify the best practices for multilingual model compression using these techniques.

\end{abstract}

\section{Introduction}
Small multilingual encoder language models (LMs), such as mBERT \cite{devlin-etal-2019-bert}, XLM-R \cite{conneau-etal-2020-unsupervised}, and Glot-500m \cite{imanigooghari-etal-2023-glot500}, have demonstrated strong performance across a diverse range of low-resource languages \cite{pmlr-v119-hu20b, asai-etal-2024-buffet}, often outperforming large-scale proprietary models on various sequential tasks \cite{adelani-etal-2024-sib, gurgurov2025smallmodelsbigimpact}. However, even these relatively compact multilingual models may still be excessively large for use in individual languages due to redundant capacity and expensive inference \cite{singh-lefever-2022-student, cruz-2025-extracting}. 

\begin{figure}[th!]
    \centering
    \includegraphics[width=0.65\linewidth]{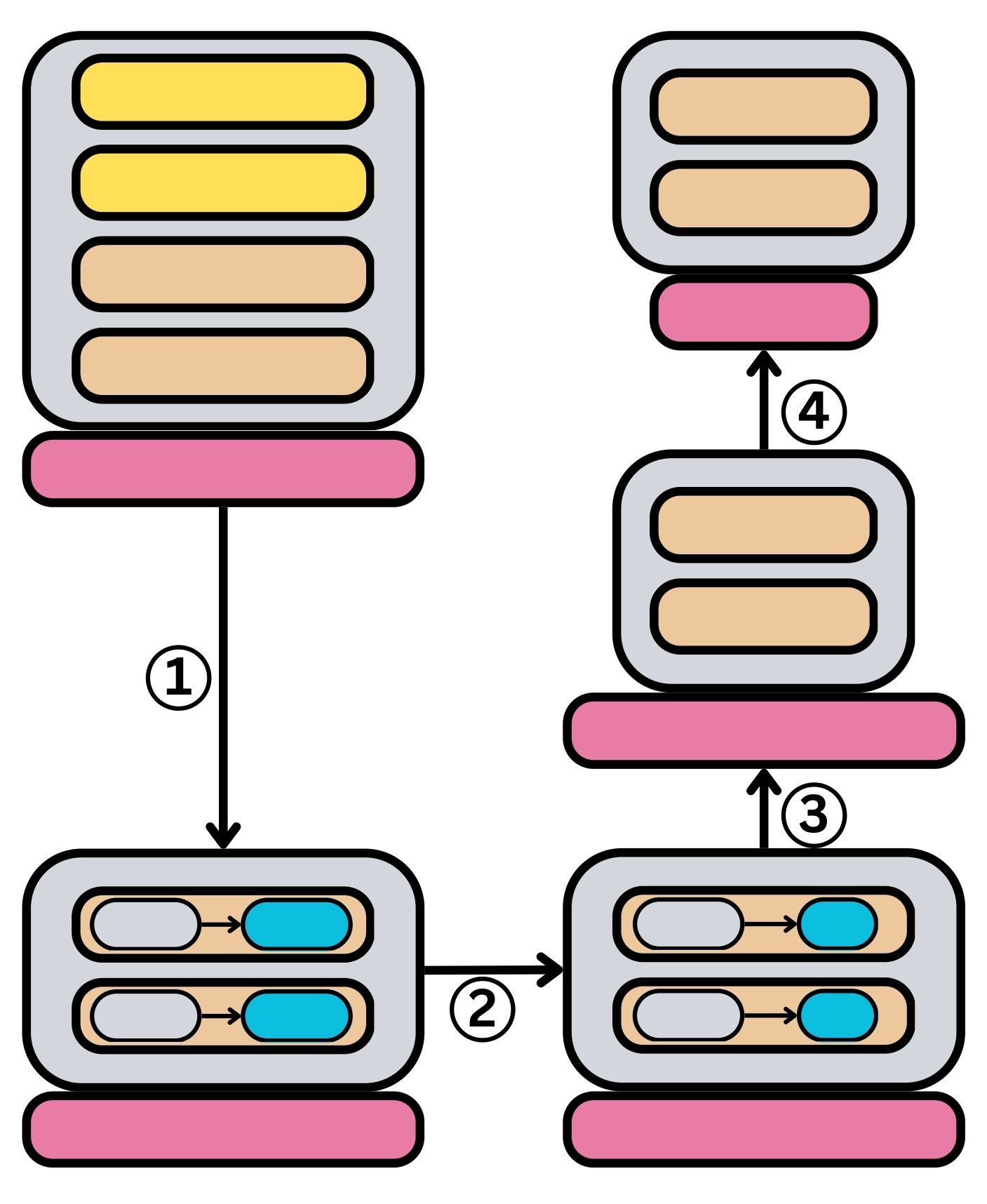}
    \caption{Overview of our multilingual model compression methodology. We use \textbf{(1)} knowledge distillation to reduce layers, \textbf{(2)} structured pruning to eliminate redundant feed-forward network width, and \textbf{(3)} hidden size reduction and another round of knowledge distillation from the previous student model. Finally, \textbf{(4)} vocabulary trimming is applied to retain language-specific tokens.}
    \label{fig:methodology}
\end{figure}

To address this, we propose a novel combination of model compression approaches for transforming multilingual encoder-only models into maximally small, efficient, language-specific alternatives while retaining competitive performance. Our methodology integrates knowledge distillation \cite{hinton2015distillingknowledgeneuralnetwork}, structured pruning \cite{kim-hassan-2020-fastformers, hou2020dynabert}, weight truncation, and vocabulary trimming \cite{abdaoui-etal-2020-load, ushio-etal-2023-efficient} to systematically reduce model size by compressing the depth (number of layers), feed-forward intermediate width, hidden size, and tokenizer vocabulary. Our experiments demonstrate that this pipeline achieves compression rates of up to 92\%, with performance drops of 2-10\% for moderate compression (up to 87\%) and 8-13\% at maximum compression on downstream tasks such as sentiment analysis, topic classification, named entity recognition, and part-of-speech tagging. Notably, for moderate compression levels, the extent of degradation depends more on the strength of the teacher model than on the compression itself. 

Beyond compression, we investigate the impact of using multilingual versus monolingual teacher models, evaluate different initialization strategies for knowledge distillation, and analyze additional compression variables. Our findings contribute to the development of highly efficient, environmentally friendly models \cite{Strubell_Ganesh_McCallum_2020} for low-resource languages and explore how strongly models can be compressed. The code for our experiments is made publicly available at \url{https://github.com/d-gurgurov/Multilingual-LM-Disitillation}. 

\section{Methodology}
In this section, we present our multilingual model compression strategy, illustrated in Figure~\ref{fig:methodology}. Our approach combines several existing compression techniques in a novel way that, to the best of our knowledge, has not been explored in this combination within the multilingual context.

\subsection{Layer Reduction via Knowledge Distillation}
We reduce the number of transformer layers in the teacher model by half to obtain an initial compact student model \cite{sanh2020distilbertdistilledversionbert}. The student is initialized with the layers of the teacher and trained using a combination of Masked Language Modeling (MLM) \cite{devlin-etal-2019-bert} and Mean Squared Error (MSE) loss for knowledge distillation \cite{hinton2015distillingknowledgeneuralnetwork} for 10 epochs. Both losses are weighted equally ($\alpha$=0.5, though other values were explored; see Appendix \ref{fig:alpha}). The teacher is a multilingual encoder fine-tuned on the target language (see Section~\ref{sec:finds}).

\subsection{Width Reduction via Structured Pruning}
We apply structured pruning \cite{kim-hassan-2020-fastformers} to reduce the intermediate size of the feed-forward layers from 3072 to 2048. Neuron importance is estimated using first-order gradient information accumulated from forward and backward passes over MLM validation data. At each layer, neurons are ranked by their absolute gradient values, and the least important ones are removed based on a target pruning ratio. The remaining neurons are then reordered to preserve model functionality. For consistency, the same pruning ratio is applied across all layers.

\subsection{Hidden Size Compression with Secondary Knowledge Distillation}
We compress the hidden embedding dimension from 768 to either 312, 456, or 564 via truncation, retaining the first $k$ dimensions.\footnote{The hidden size must be divisible by the number of attention heads.} A second round of knowledge distillation is then performed, using the width-reduced model from the previous step as the new teacher, similar to \citet{10254426}, with training for 10 epochs.

\subsection{Vocabulary Reduction}
We reduce the vocabulary size by selecting the top 40,000 most frequent tokens from a target-language corpus, along with their corresponding embeddings \cite{ushio-etal-2023-efficient}. This ensures that the resulting model retains only language-specific tokens, which significantly reduces the overall model size.

\section{Experiments}

Below, we describe the datasets, languages, tasks, and baseline systems used in our evaluation.

\subsection{Knowledge Distillation Data}
We use GlotCC \cite{kargaran2025glotccopenbroadcoveragecommoncrawl}, a large-scale multilingual corpus derived mainly from CommonCrawl \cite{wenzek-etal-2020-ccnet}, as the primary dataset for both stages of knowledge distillation. Data distributions for the selected languages are reported in Appendix~\ref{app:data}. We use GlotCC for training, and the FLORES-200 development set \cite{nllbteam2022languageleftbehindscaling} for validation during training.


\subsection{Languages and Tasks}


We evaluate our models on four tasks: Topic Classification (TC), Sentiment Analysis (SA), Named Entity Recognition (NER), and Part-of-Speech Tagging (POS), covering three low-resource languages--Maltese, Slovak, and Swahili \cite{joshi2020state}. For TC, we use the 7-class SIB-200 dataset \cite{adelani-etal-2024-sib}, and for SA, we compile binary sentiment datasets from multiple sources \cite{dingli2016sentiment, cortis-davis-2019-social, pecar-etal-2019-improving, Muhammad2023AfriSentiAT, muhammad2023semeval}. For NER, we use WikiANN \cite{pan-etal-2017-cross}, and for POS, we use Universal Dependencies v2.15 \cite{de-marneffe-etal-2021-universal} and MasakhaPOS \cite{dione-etal-2023-masakhapos}. For all tasks, we train Sequential Bottleneck task adapters \cite{pfeiffer-etal-2020-mad} with fixed hyperparameters (see Appendix \ref{app:hyper}). Performance is measured using macro-averaged F1 \cite{sokolova2006beyond} for TC and SA, and "seqeval" F1 \cite{seqeval} for NER and POS.

\subsection{Models and Baselines}
We compress two encoder multilingual models--mBERT \cite{devlin-etal-2019-bert} and XLM-R-base \cite{conneau-etal-2020-unsupervised}--adapted to target languages through fine-tuning on language-specific data, and compare the reduced models to two baselines: (1) the original, non-adapted models, and (2) language-adapted versions. In both cases, we train an identical task adapter using the same task-specific datasets as for the compressed models. 

\section{Findings}
\label{sec:finds}
Our key findings are outlined below.


\subsection{Distillation} 

\begin{figure}[h]
    \centering
    \begin{subfigure}{0.48\linewidth}
        \centering
        \includegraphics[width=\linewidth]{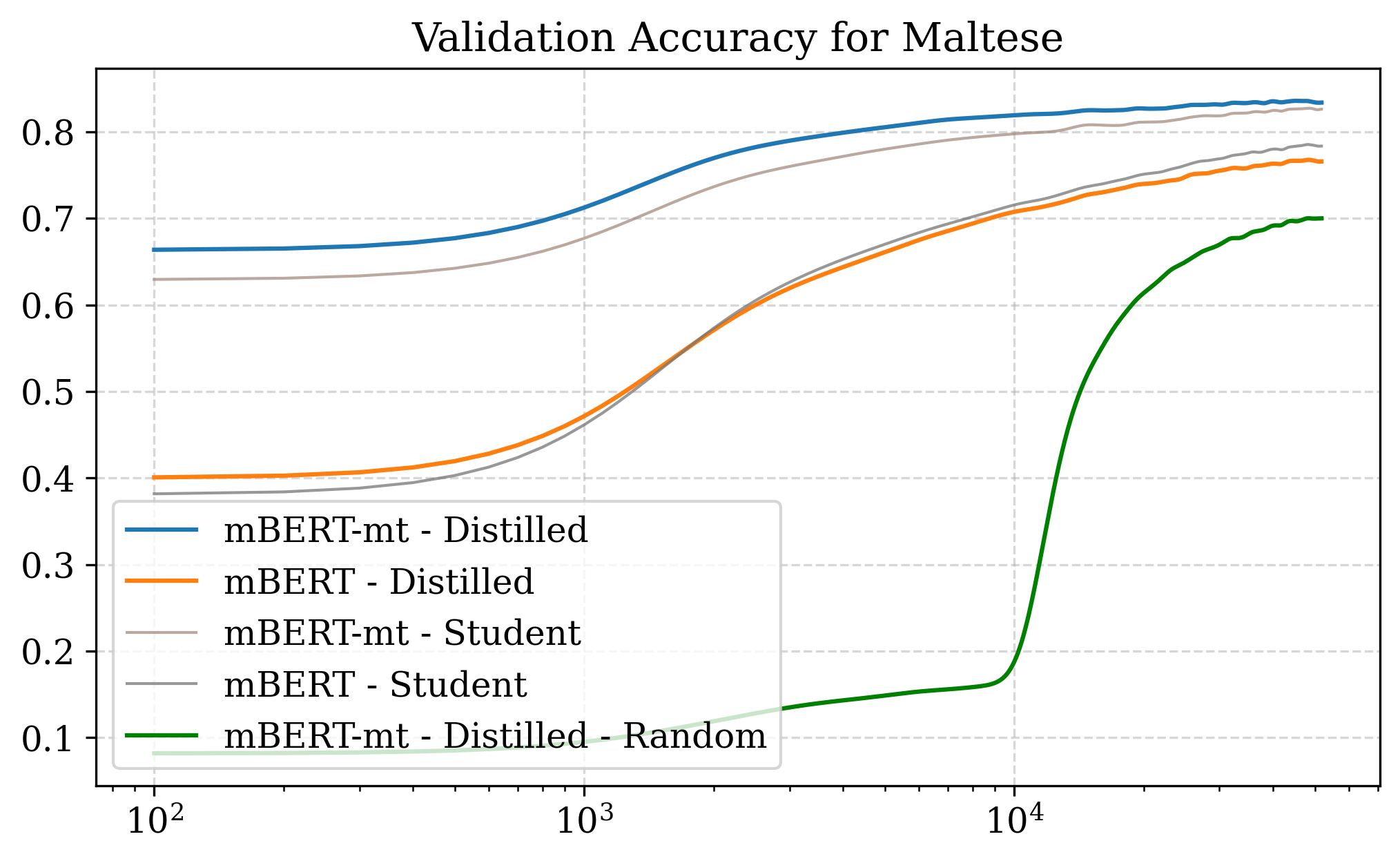}
        \caption{Maltese (mBERT)}
        \label{fig:kd_1_mt}
    \end{subfigure}
    \hfill
    \begin{subfigure}{0.48\linewidth}
        \centering
        \includegraphics[width=\linewidth]{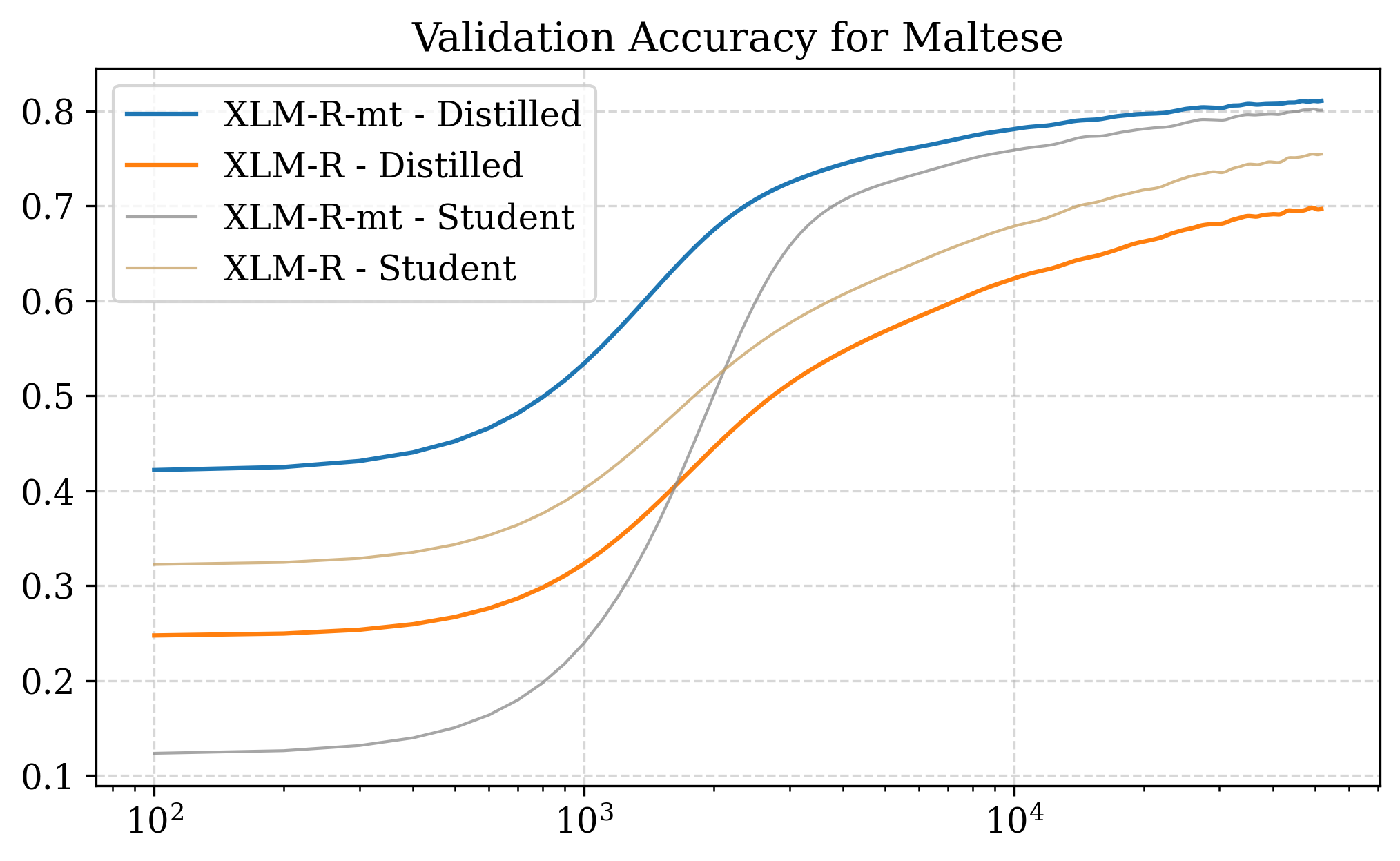}
        \caption{Maltese (XLM-R)}
        \label{fig:kd_1_mt_xlm}
    \end{subfigure}

    \vspace{0.5cm}

    \begin{subfigure}{0.48\linewidth}
        \centering
        \includegraphics[width=\linewidth]{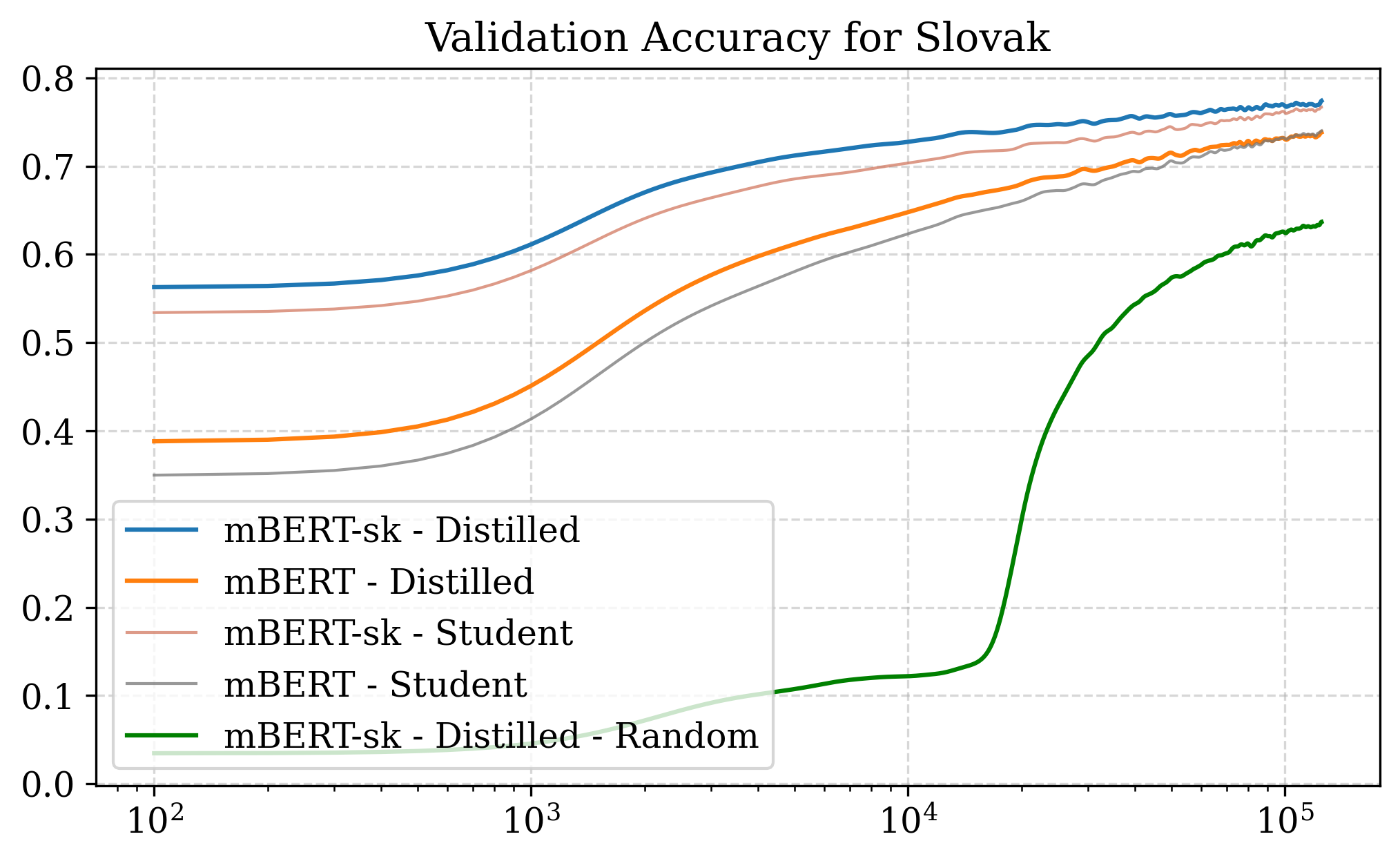}
        \caption{Slovak (mBERT)}
        \label{fig:kd_1_sk}
    \end{subfigure}
    \hfill
    \begin{subfigure}{0.48\linewidth}
        \centering
        \includegraphics[width=\linewidth]{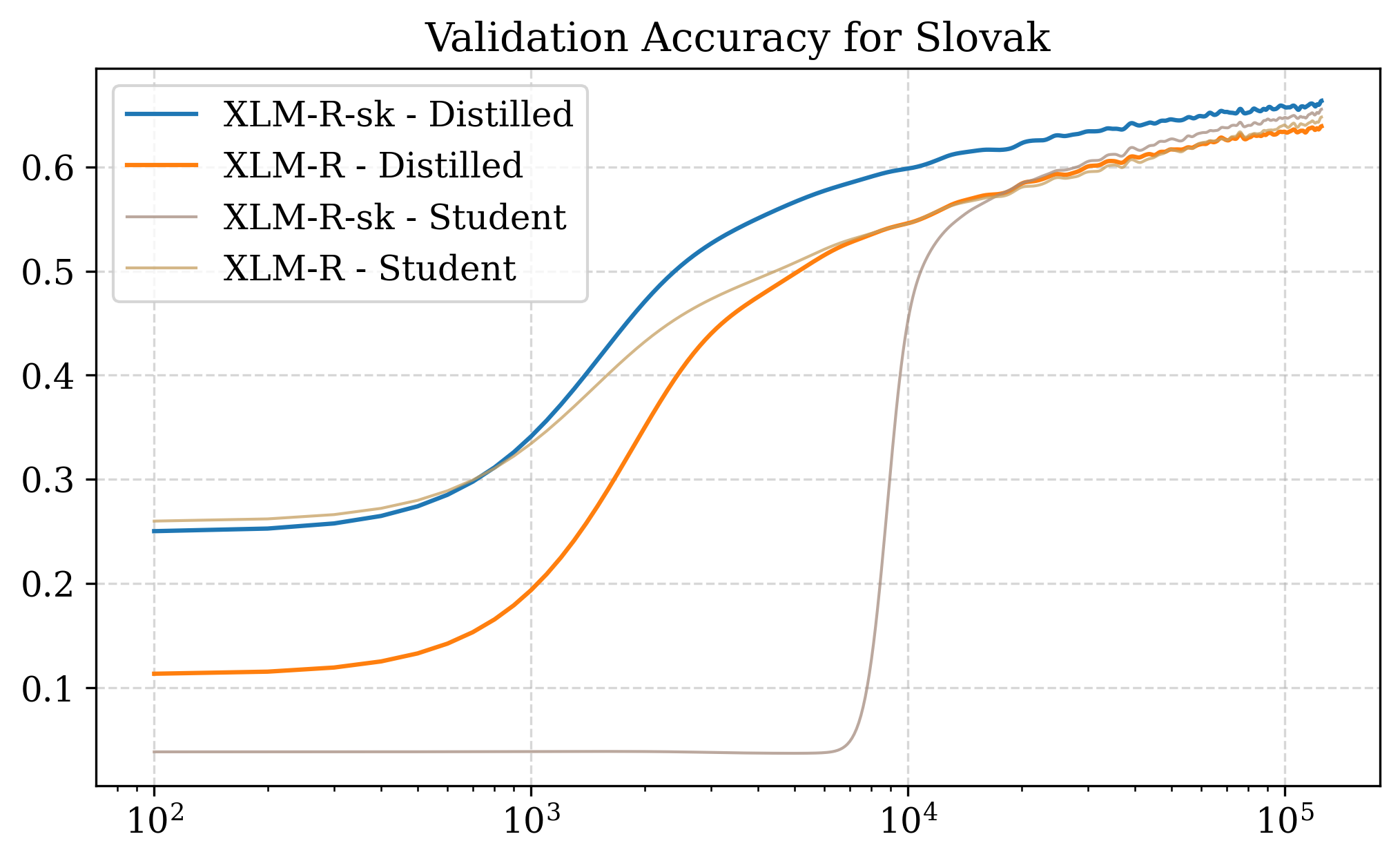}
        \caption{Slovak (XLM-R)}
        \label{fig:kd_1_sk_xlm}
    \end{subfigure}

    \vspace{0.5cm}

    \begin{subfigure}{0.48\linewidth}
        \centering
        \includegraphics[width=\linewidth]{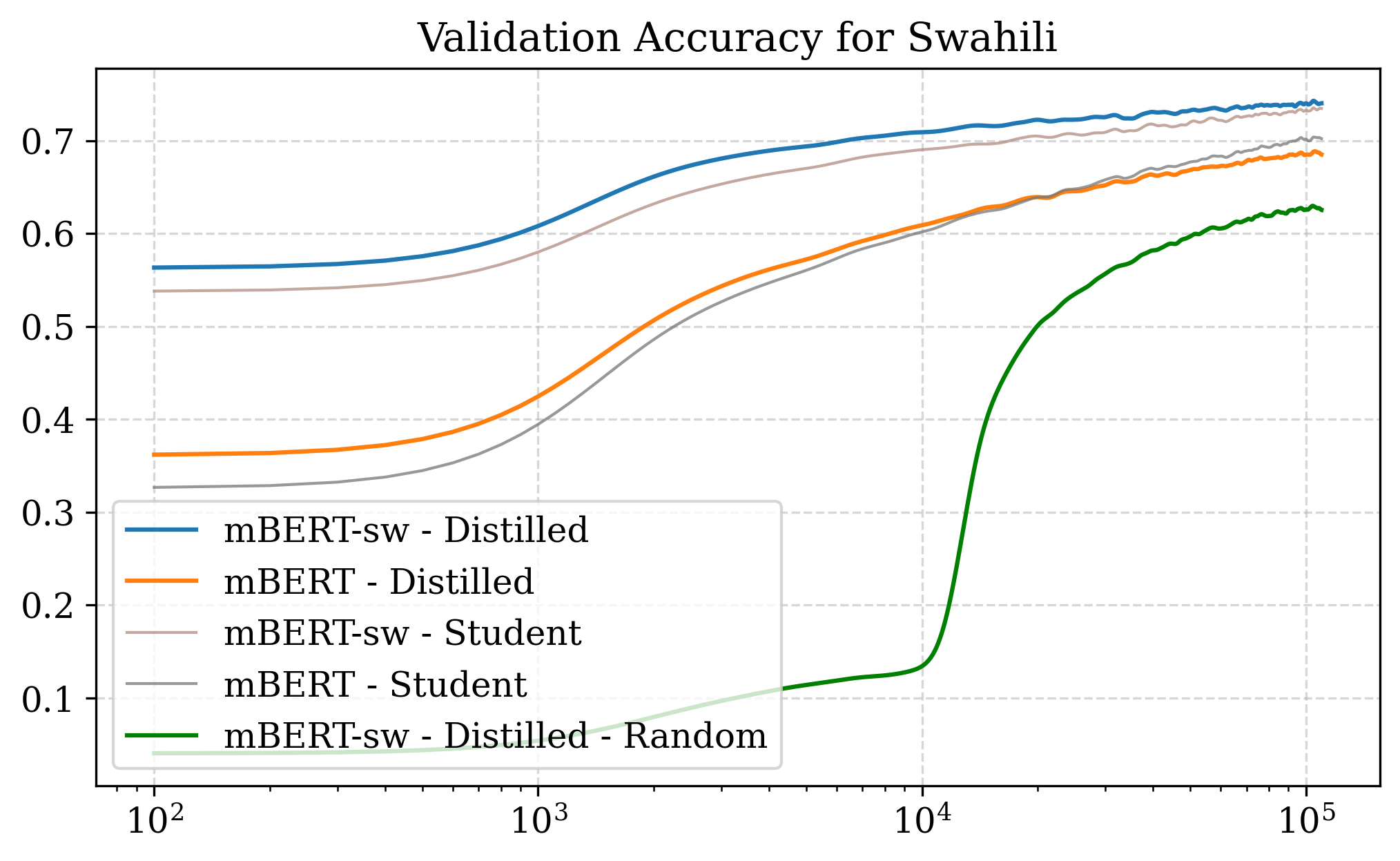}
        \caption{Swahili (mBERT)}
        \label{fig:kd_1_sw}
    \end{subfigure}
    \hfill
    \begin{subfigure}{0.48\linewidth}
        \centering
        \includegraphics[width=\linewidth]{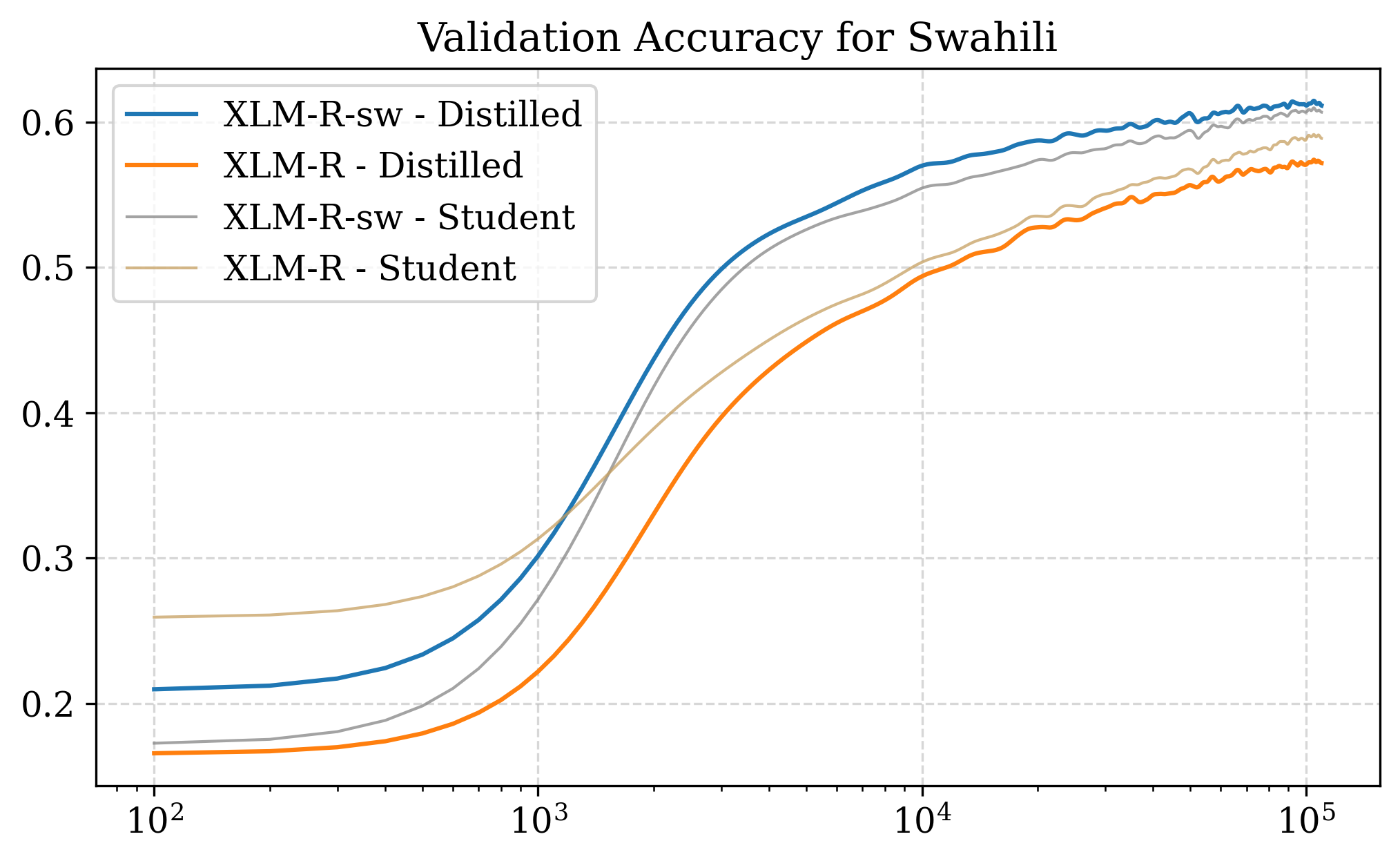}
        \caption{Swahili (XLM-R)}
        \label{fig:kd_1_sw_xlm}
    \end{subfigure}

    \caption{First-step KD validation accuracies for mBERT and XLM-R with models initialized using the last $k$ layers. mBERT- and XLM-R-{mt, sk, sw} refer to models adapted to the target language; \textit{distilled} denotes models trained with distillation loss, while \textit{student} refers to identically trained models without distillation loss. The best accuracy is in all cases achieved when distilling from a target-language adapted model.}
    \label{fig:kd_1_combined}
\end{figure}

Distilling knowledge from a multilingual teacher into a monolingual student model is less effective than using a target-language adapted teacher, as evidenced by the differences in validation accuracies shown in Figure \ref{fig:kd_1_combined}. This discrepancy possibly stems from the multilingual teacher's broad cross-lingual representations, which are not directly aligned with the requirements of a monolingual student. In contrast, monolingual teachers provide more targeted, language-specific representations, resulting in better student performance.

\textbf{Distillation loss:} We compare KL divergence and MSE as distillation loss functions, and observe that MSE leads to better and faster convergence (Appendix~\ref{app:kl_mse}), in line with prior work \cite{kim2021comparingkullbackleiblerdivergencemean, nityasya2022studentbestcomprehensiveknowledge}.



\subsection{Weight Initialization} 

Weight initialization plays a crucial role in training the student model, with knowledge distillation providing only a marginal additional performance improvement (Figure \ref{fig:kd_1_combined}). This partly aligns with the findings of \citet{wibowo2024privilegedstudentsvalueinitialization}, who explored distilling multilingual abilities for multilingual tasks, whereas our focus is on monolingual distillation. Training a student-sized model initialized with teacher weights, but without knowledge distillation, results in a slight performance drop compared to a fully distilled model. 

\textbf{Initialization Strategies:} Among various initialization strategies, initializing the student with the last $k$ layers for mBERT and every other layer (stride) for XLM-R consistently outperforms alternatives such as using the first $k$ layers and combining first and last layers (Appendix~\ref{app:init_kd}). Random initialization performs significantly worse, emphasizing the importance of weight reuse \cite{sun2019patientknowledgedistillationbert, singh-lefever-2022-student}.


\begin{table*}[h]
\centering
\small
\resizebox{\textwidth}{!}{
\begin{tabular}{@{}l r r  @{\hspace{0.6em}}rrrr rrrr rrrr r@{}} 
\toprule
\multirow{2}{*}{\textbf{Compression Stage}} & \multirow{2}{*}{\textbf{Params}} & \multirow{2}{*}{\textbf{Size}} & \multicolumn{12}{c}{\textbf{Task Performance (F1)}} & \multirow{2}{*}{\textbf{Avg}} \\
\cmidrule(lr){4-15}
& & & \multicolumn{4}{c}{\textbf{Maltese}} & \multicolumn{4}{c}{\textbf{Slovak}} & \multicolumn{4}{c}{\textbf{Swahili}} \\ 
\cmidrule(lr){4-7} \cmidrule(lr){8-11} \cmidrule(lr){12-15}
& & & TC & SA & NER & POS & TC & SA & NER & POS & TC & SA & NER & POS \\ 
\midrule
\multicolumn{15}{l}{\textit{Baselines}} \\
\quad Multilingual & 279M & 279M & 68.1 & 56.0 & 54.3 & 89.9 & 88.1 & 95.6 & 91.1 & 97.3 & 78.4 & 81.5 & 84.6 & 89.4 & 81.2 \\ 
\rowcolor{gray!10}
\quad Language-adapted & 279M & 279M & 85.0 & 76.2 & 69.2 & 95.4 & 86.2 & 94.8 & 91.0 & 97.1 & 87.5 & 84.1 & 82.7 & 89.2 & \textbf{86.5} \\ 
\midrule
\multicolumn{15}{l}{\textit{Compression Pipeline (minimal degradation)}} \\
\quad Layer reduction & 236M (-15\%) & 236M & 84.0 & 77.2 & 63.5 & 94.3 & 86.3 & 92.9 & 90.1 & 96.3 & 82.9 & 81.3 & 82.9 & 89.2 & 85.1 \\ 
\quad + FFN pruning & 226M (-20\%) & 226M & 84.7 & 78.6 & 60.1 & 94.2 & 86.1 & 93.4 & 90.0 & 96.1 & 82.4 & 82.7 & 83.6 & 89.5 & 85.1 \\ 
\quad + Hidden 564 & 163M (-40\%) & 163M & 83.4 & 74.9 & 53.0 & 93.7 & 84.9 & 92.7 & 89.1 & 96.8 & 85.8 & 81.0 & 80.8 & 89.4 & 83.8 \\ 
\rowcolor{green!20}
\quad + Vocabulary & 45M (-85\%) & 45M & 84.1 & 72.4 & 60.9 & 93.0 & 85.3 & 92.9 & 89.3 & 96.4 & 85.7 & 80.9 & 82.0 & 89.1 & \textbf{84.3} \\ 
\midrule
\multicolumn{15}{l}{\textit{Further compression (moderate degradation)}} \\
\quad + Hidden 456 & 131M (-53\%) & 131M & 78.5 & 69.9 & 62.5 & 92.7 & 86.0 & 93.0 & 88.3 & 96.3 & 83.1 & 79.3 & 80.7 & 88.9 & 83.3 \\ 
\quad + Vocabulary & 35M (-87\%) & 35M & 78.5 & 70.7 & 63.3 & 92.5 & 86.1 & 92.9 & 88.4 & 96.3 & 82.5 & 79.0 & 80.2 & 89.0 & 83.3 \\ 
\midrule
\multicolumn{15}{l}{\textit{Maximum compression (higher degradation)}} \\
\rowcolor{orange!15}
\quad + Hidden 312 & 89M (-68\%) & 89M & 66.9 & 70.1 & 35.7 & 87.6 & 84.0 & 90.9 & 88.0 & 95.5 & 76.4 & 80.1 & 80.7 & 88.3 & 78.7 \\ 
\rowcolor{orange!15}
\quad + Vocabulary & 23M (-92\%) & 23M & 67.2 & 71.4 & 37.1 & 87.5 & 84.0 & 90.5 & 88.2 & 95.6 & 78.0 & 80.5 & 79.2 & 88.0 & 78.9 \\ 
\bottomrule
\end{tabular}
}
\caption{Progressive compression of XLM-R-base. Stages are grouped by degradation level. Highlighted rows indicate the baseline (gray) and optimal compression point (green, 85\% reduction with 2.5\% drop). Maximum compression rows (red) show higher degradation rates (7.6\% drop). All F1 scores are averaged over 3 independent runs with different random seeds mBERT in Appendix~\ref{app:mbert}.}
\label{tab:performance}
\end{table*}

\subsection{Pruning and Truncation} 

Distilled models can be compressed further using structured pruning, hidden size reduction, and vocabulary trimming, while maintaining competitive performance.

\textbf{Intermediate size reduction:} Reducing the intermediate size of feed-forward layers from 3072 to 2048 via structured pruning results in negligible performance loss (Table~\ref{tab:performance}). However, more aggressive reductions degrade quality significantly, making 2048 a practical lower bound. We do not prune attention heads, as removing even a minimal number (e.g., three) causes severe degradation (>50\% performance drop in preliminary experiments).

\textbf{Hidden size reduction:} We reduce the hidden embedding size to 564, 456, and 312, truncating it to the first $k$ dimensions. Training is performed under the supervision of the student from the previous stage. We find that using the original teacher leads to worse results, possibly due to the bigger knowledge gap \cite{10254426}. We also tested SVD-based dimensionality reduction but found truncation to be more effective (see Appendix \ref{app:svd}).


\textbf{Vocabulary trimming:} Restricting the vocabulary to the top 40K most frequent tokens for each target language introduces no measurable performance loss compared to the previous step, while further improving efficiency. Reducing below 40K works for some languages but does not generalize well across all cases (Appendix \ref{app:vocab}), consistent with \citet{ushio-etal-2023-efficient}.

\subsection{Downstream Performance}  

Our results show that model compression through knowledge distillation, structured pruning, and vocabulary reduction leads to modest performance drops (Tables \ref{tab:performance} and \ref{tab:performance-mbert}). Below, we report results for XLM-R; results for mBERT follow similar patterns and are presented in Appendix \ref{app:mbert}. 

\textbf{Language-specific resilience:} The extent of degradation varies by language and correlates with teacher model quality. At maximum compression (92\% parameter reduction), Slovak (1032MB fine-tuning (FT) data) experiences only a 2.9\% performance drop, Swahili (332MB) shows a 5.2\% drop, while Maltese (188MB) degrades by 19.2\%. This pattern demonstrates that stronger teacher models--trained on larger datasets--enable more robust compression outcomes.

\textbf{Task-specific patterns:} Different tasks exhibit varying compression sensitivities. POS tagging shows the highest resilience across all languages, with performance drops of only 4-13\% at 92\% compression. Conversely, NER demonstrates steeper degradation, particularly for Maltese (69.2 → 37.1 F1). This severe drop is likely compounded by the extremely small Maltese NER training set (100 examples vs. 20,000 for Slovak), indicating that sequence labeling tasks are especially vulnerable to compression in low-resource settings. In contrast, sentence-level classification tasks such as SA and TC remain relatively stable under heavy compression, with performance decreases below 10\% even at 85–90\% size reduction.

\textbf{Optimal compression trade-offs:} The 85\% compression level (hidden size 564 with 40k vocabulary) offers the best balance for most scenarios, with only a 2.5\% average performance drop (84.3 vs 86.5 avg F1). For high-resource languages like Slovak, even 87\% compression incurs only a 3.8\% drop. Notably, vocabulary trimming often yields slight improvements (e.g., Maltese TC: 84.11 vs 83.43 F1), suggesting it reduces vocabulary noise while compensating for hidden size reduction.

\textbf{Staged compression effects:} Layer reduction (15\%) and intermediate size pruning (20\%) induce minimal degradation (<2\% drop), with the primary performance impact occurring during hidden size reduction. Performance degrades gradually up to 85\% compression, but deteriorates more rapidly beyond this threshold (4-6\% drop per additional stage).

\textbf{Adapter capacity:} We experiment with varying the reduction factor $r$ to adjust task adapter capacity (Appendix~\ref{app:red_factor}, Figure~\ref{fig:red}). While $r=16$ suffices for larger models, smaller models (hidden sizes 564, 456, 312) benefit from lower $r$ values ($r=2$), yielding modest performance gains. Results in Tables~\ref{tab:performance} and~\ref{tab:performance-mbert} use $r=2$ for these compressed models.

\section{Related Work}

In knowledge distillation, a smaller student model is trained to replicate the behavior of a larger teacher model \cite{hinton2015distillingknowledgeneuralnetwork}, often combining MLM loss with teacher supervision \cite{sun2019patientknowledgedistillationbert, sanh2020distilbertdistilledversionbert}. DistilBERT \cite{sanh2020distilbertdistilledversionbert} reduces model size by selecting every other layer from BERT \cite{devlin-etal-2019-bert} and distills on large corpora using dynamic masking. Patient distillation further improves results by matching intermediate representations \cite{sun2019patientknowledgedistillationbert}.


Recent work has explored distilling multilingual models into compact monolingual models. \citet{singh-lefever-2022-student} train student models for languages such as Swahili and Slovenian using a composite loss (distillation, cosine, MLM), and show that distilled models often outperform mBERT while using a reduced vocabulary \cite{abdaoui-etal-2020-load}. \citet{ansell2023distillingefficientlanguagespecificmodels} introduce a two-phase bilingual distillation pipeline, combining general-purpose and task-specific guidance with sparse fine-tuning, outperforming multilingual baselines.

Other studies emphasize the role of initialization. \citet{wibowo2024privilegedstudentsvalueinitialization} show that copying teacher weights is more effective than random initialization in the context of multilingual distillation, and that MSE outperforms KL divergence for distillation. \citet{cruz-2025-extracting} similarly distill mBERT for Tagalog and highlight the nuanced impact of embedding initialization.

\section{Conclusion}
We present an effective compression pipeline for multilingual encoder models designed for low-resource languages. By integrating staged knowledge distillation, structured pruning, hidden size truncation, and vocabulary reduction, we compress models by up to 92\% while maintaining competitive performance, typically within 2–10\% of the original for moderate compression and 8–13\% at maximum compression, on four downstream tasks.



\section*{Limitations}

Our evaluation is limited to three low-resource languages and four downstream tasks, which may affect generalizability to other languages and task types. The compression pipeline requires target-language data for teacher adaptation, making it less suitable for truly low-resource languages with minimal corpora. We focus exclusively on encoder-only models (mBERT and XLM-R), and our structured pruning only targets feed-forward layers, leaving attention head pruning unexplored due to performance degradation.


\section*{Acknowledgments}
This research was supported by DisAI - Improving scientific excellence and creativity in combating disinformation with artificial intelligence and language technologies, a Horizon Europe-funded project under GA No. 101079164, by the German Ministry of Education and Research (BMBF) as part of the project TRAILS (01IW24005), and by lorAI - Low Resource Artificial Intelligence, a project funded by the European Union under GA No.101136646.

\bibliography{custom}
\bibliographystyle{acl_natbib}

\newpage
\appendix

\section{KL Divergence vs MSE for Knowledge Distillation}
\label{app:kl_mse}

\begin{figure}[h]
    \centering
    \begin{subfigure}{0.48\linewidth}
        \centering
        \includegraphics[width=\linewidth]{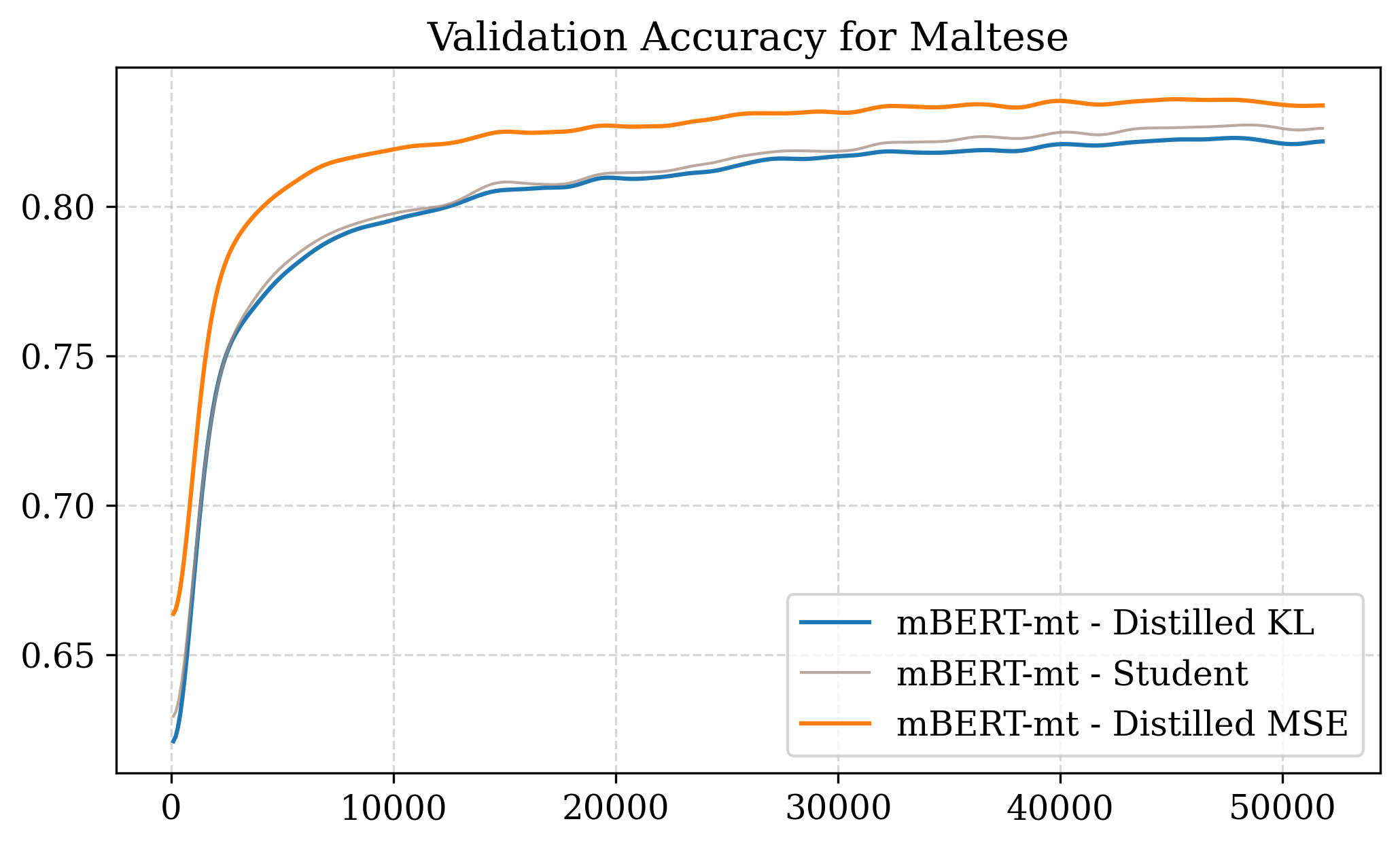}
        \caption{Maltese}
        \label{fig:mse_kd_mt}
    \end{subfigure}
    \hfill
    \begin{subfigure}{0.48\linewidth}
        \centering
        \includegraphics[width=\linewidth]{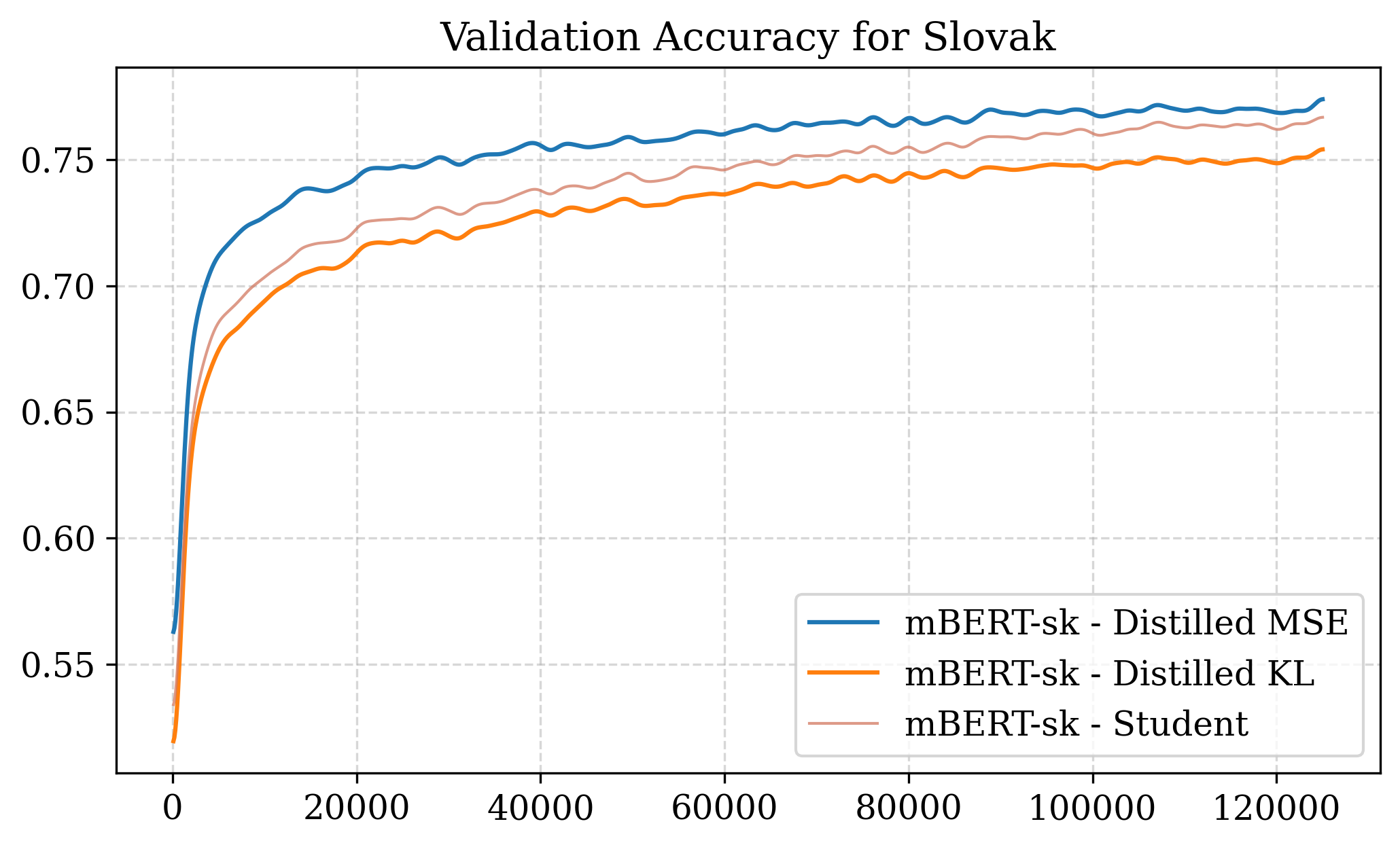}
        \caption{Slovak}
        \label{fig:mse_kd_sk}
    \end{subfigure}
    
    \vspace{0.5cm} 

    \begin{subfigure}{0.48\linewidth}
        \centering
        \includegraphics[width=\linewidth]{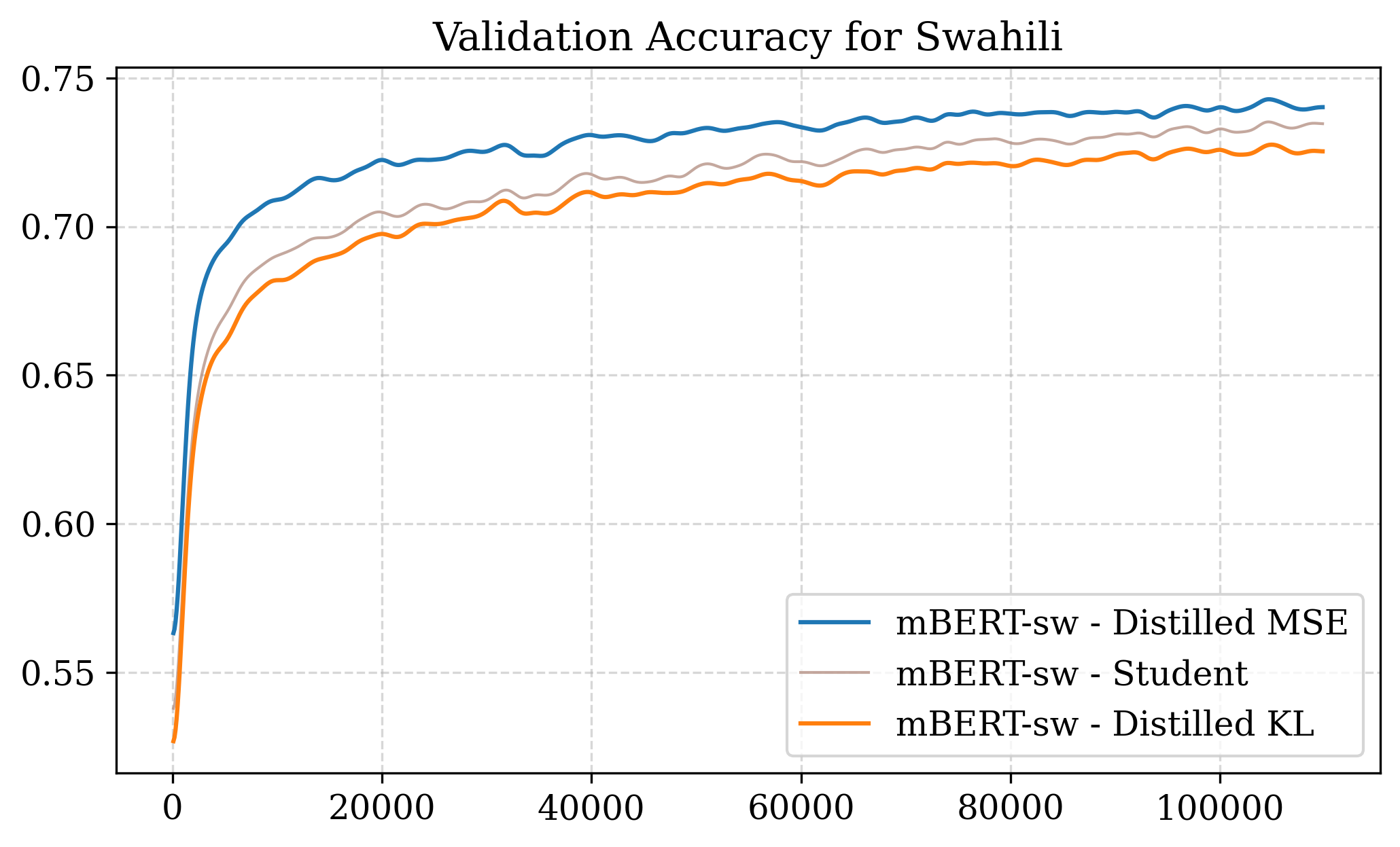}
        \caption{Swahili}
        \label{fig:mse_kd_sw}
    \end{subfigure}

    \caption{MSE vs. KD validation accuracy for mBERT with the models initialized using the last $k$ layers.}
    \label{fig:mse_kd}
\end{figure}

\begin{figure}[h]
    \centering
    \begin{subfigure}{0.48\linewidth}
        \centering
        \includegraphics[width=\linewidth]{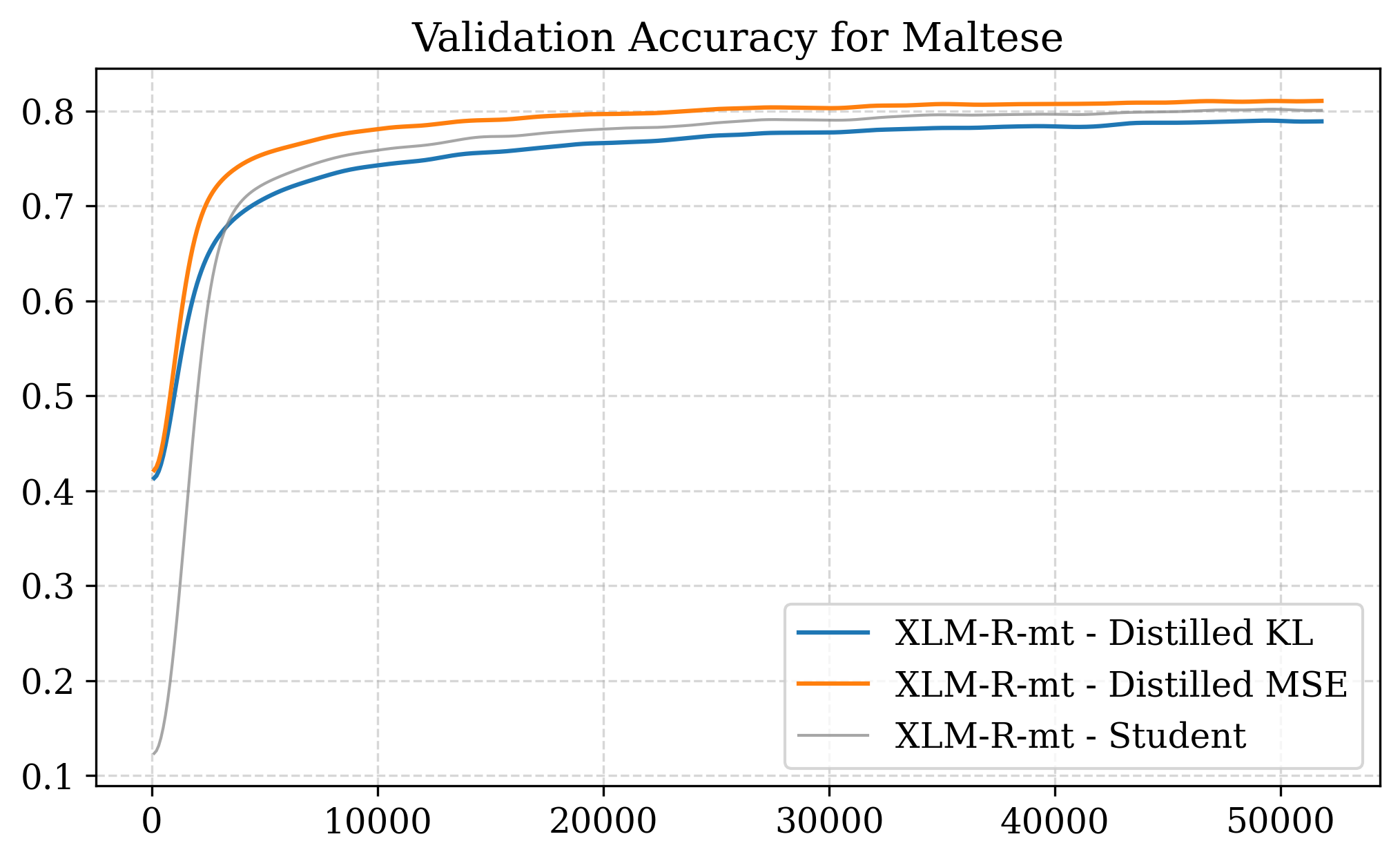}
        \caption{Maltese}
        \label{fig:mse_kd_mt_xlm}
    \end{subfigure}
    \hfill
    \begin{subfigure}{0.48\linewidth}
        \centering
        \includegraphics[width=\linewidth]{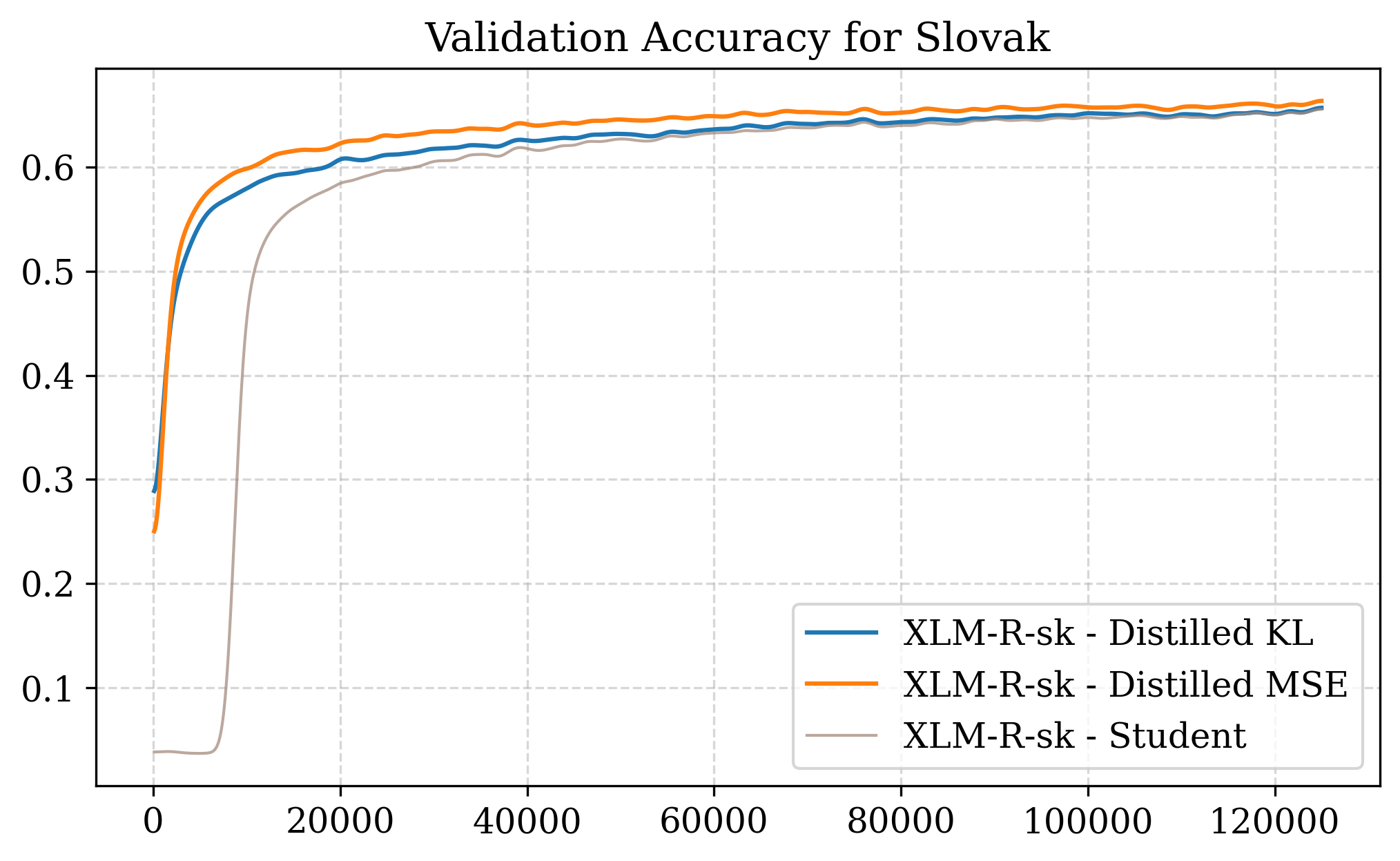}
        \caption{Slovak}
        \label{fig:mse_kd_sk_xlm}
    \end{subfigure}
    
    \vspace{0.5cm} 

    \begin{subfigure}{0.48\linewidth}
        \centering
        \includegraphics[width=\linewidth]{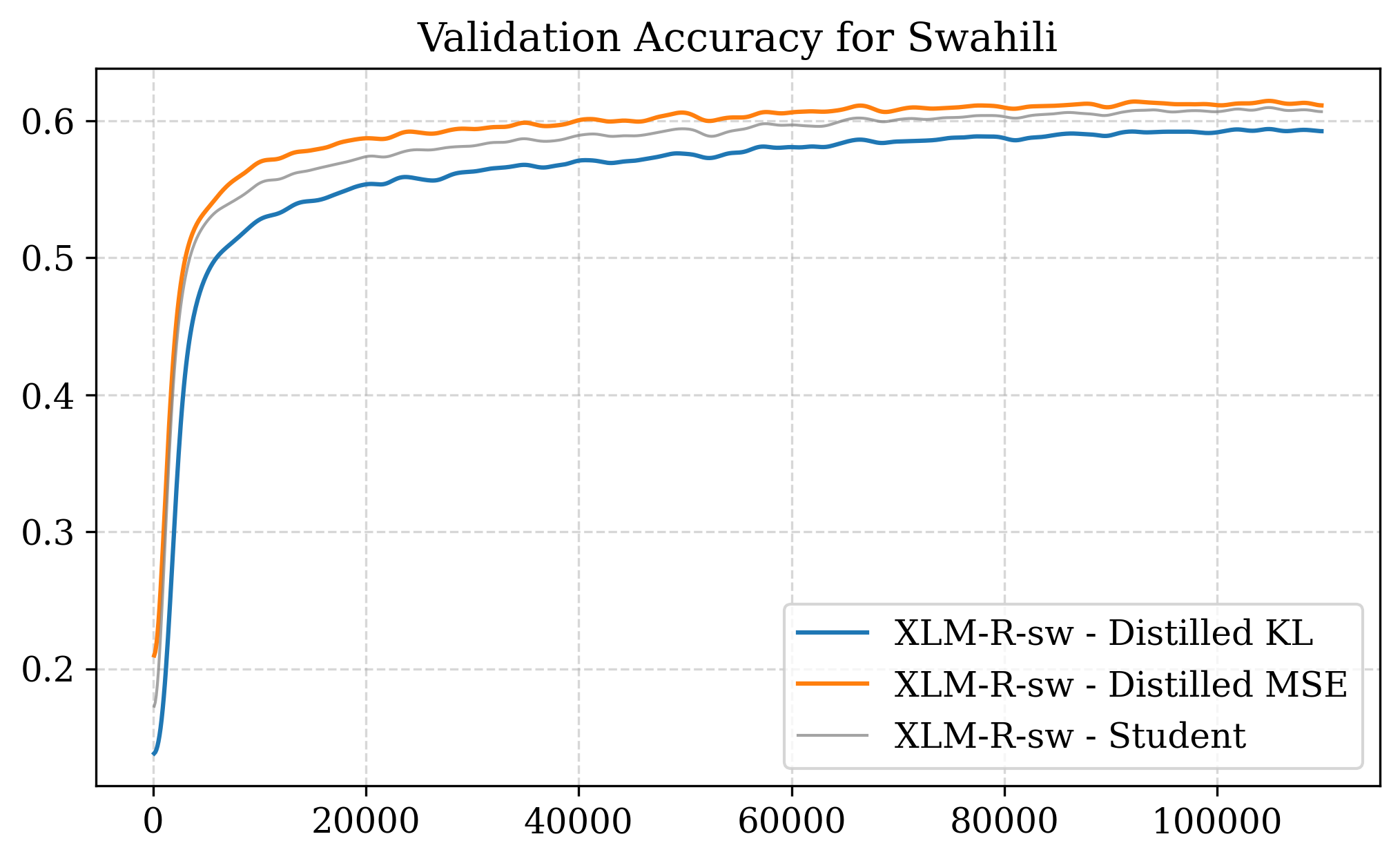}
        \caption{Swahili}
        \label{fig:mse_kd_sw_xlm}
    \end{subfigure}

    \caption{MSE vs. KD validation accuracy for XLM-R with the models initialized using the last $k$ layers.}
    \label{fig:mse_kd_xlm}
\end{figure}

\newpage
\section{Initialization Strategies for Knowledge Distillation}
\label{app:init_kd}

\begin{figure}[h]
    \centering
    \begin{subfigure}{0.48\linewidth}
        \centering
        \includegraphics[width=\linewidth]{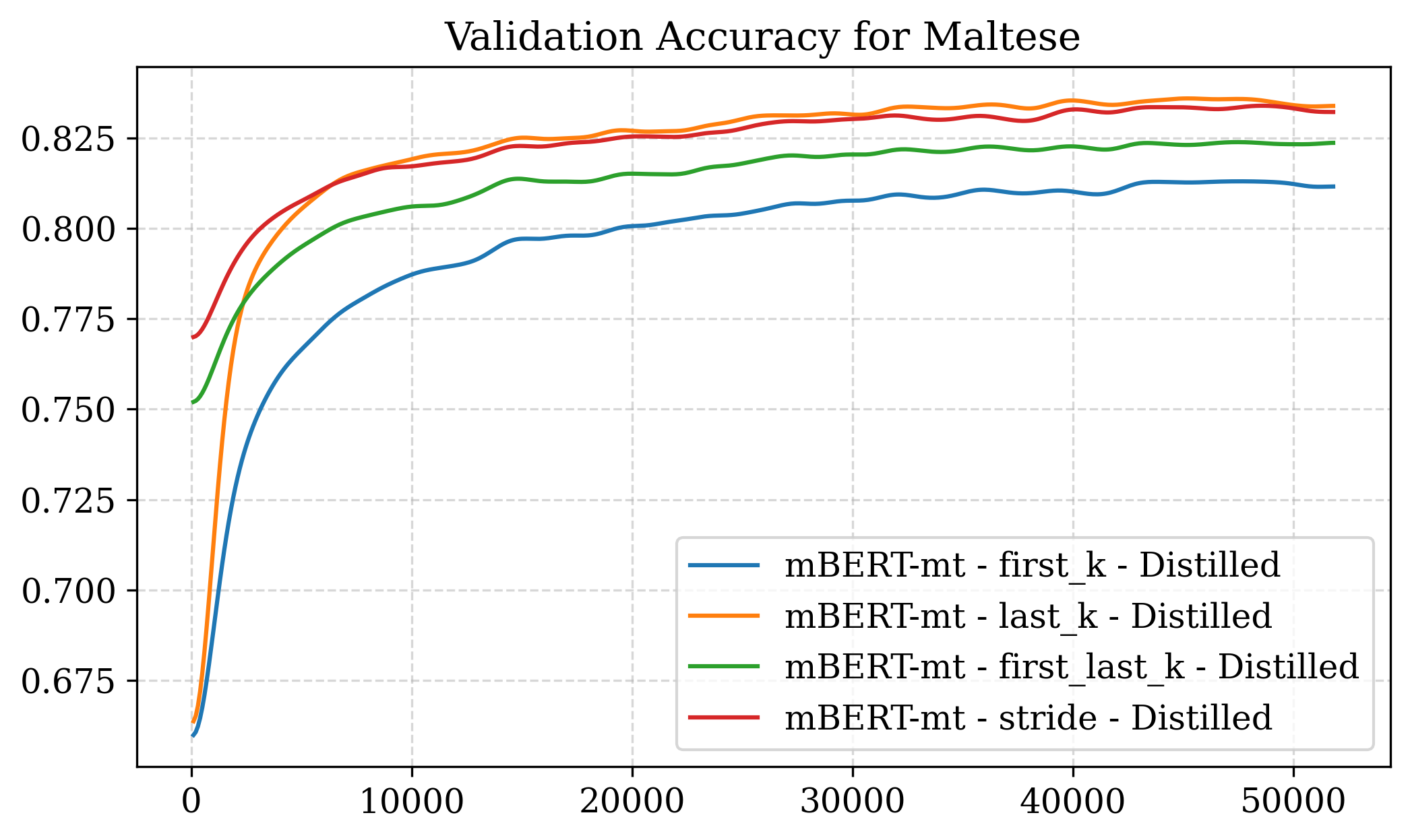}
        \caption{Maltese}
        \label{fig:init_kd_mt}
    \end{subfigure}
    \hfill
    \begin{subfigure}{0.48\linewidth}
        \centering
        \includegraphics[width=\linewidth]{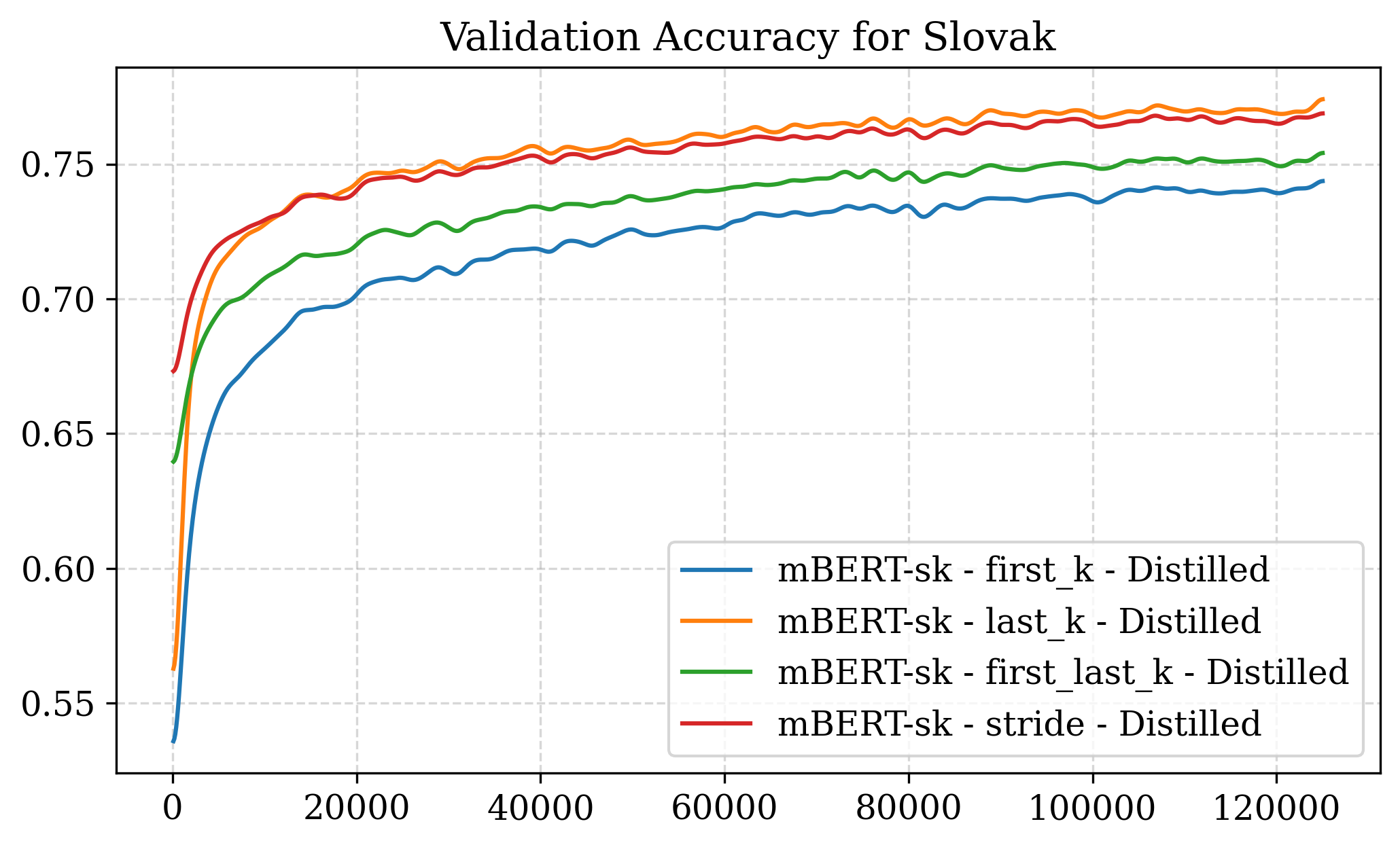}
        \caption{Slovak}
        \label{fig:init_kd_sk}
    \end{subfigure}
    
    \vspace{0.5cm} 

    \begin{subfigure}{0.48\linewidth}
        \centering
        \includegraphics[width=\linewidth]{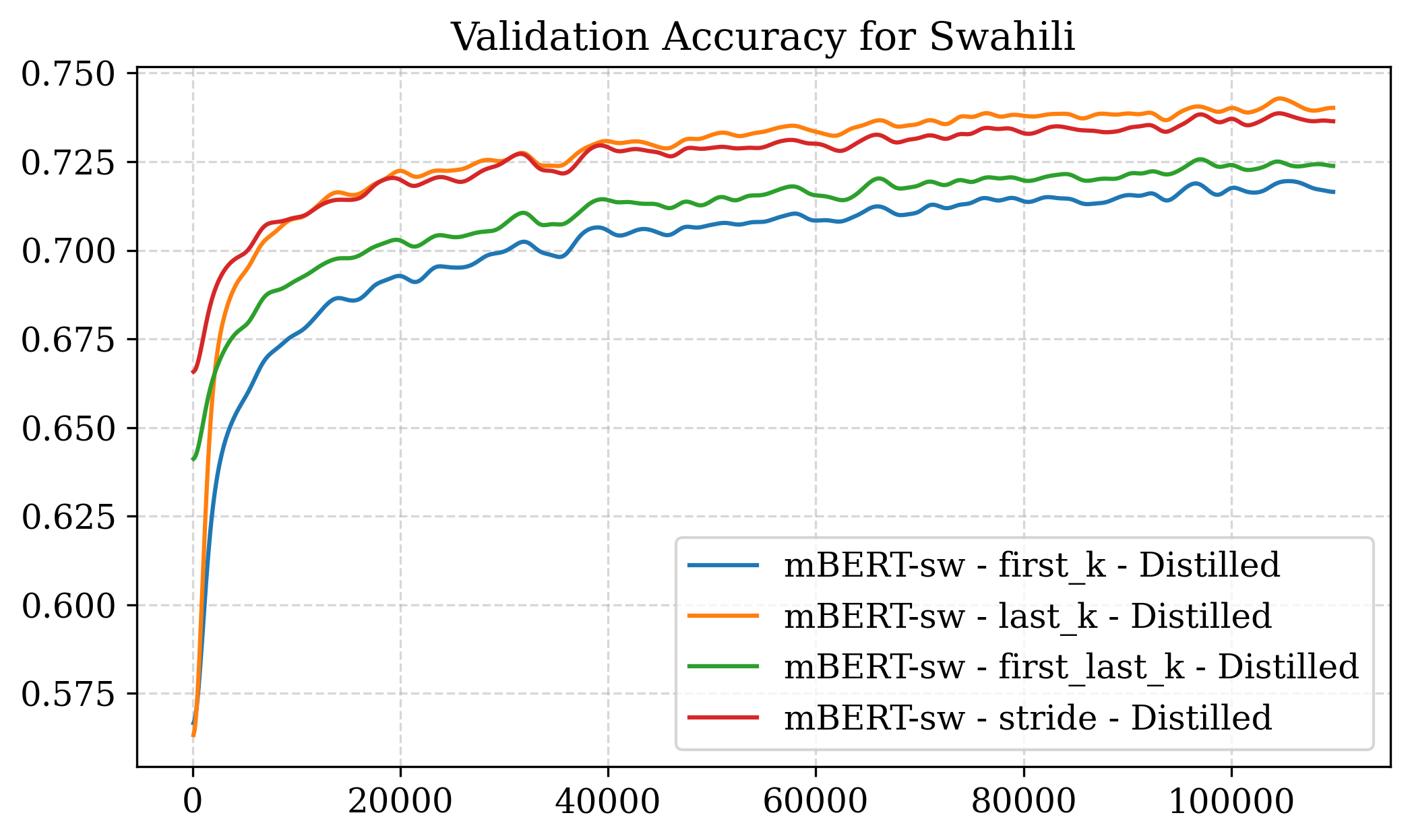}
        \caption{Swahili}
        \label{fig:init_kd_sw}
    \end{subfigure}

    \caption{Validation accuracy for various initialization strategies for mBERT.}
    \label{fig:init_kd_mbert}
\end{figure}

\begin{figure}[h]
    \centering
    \begin{subfigure}{0.48\linewidth}
        \centering
        \includegraphics[width=\linewidth]{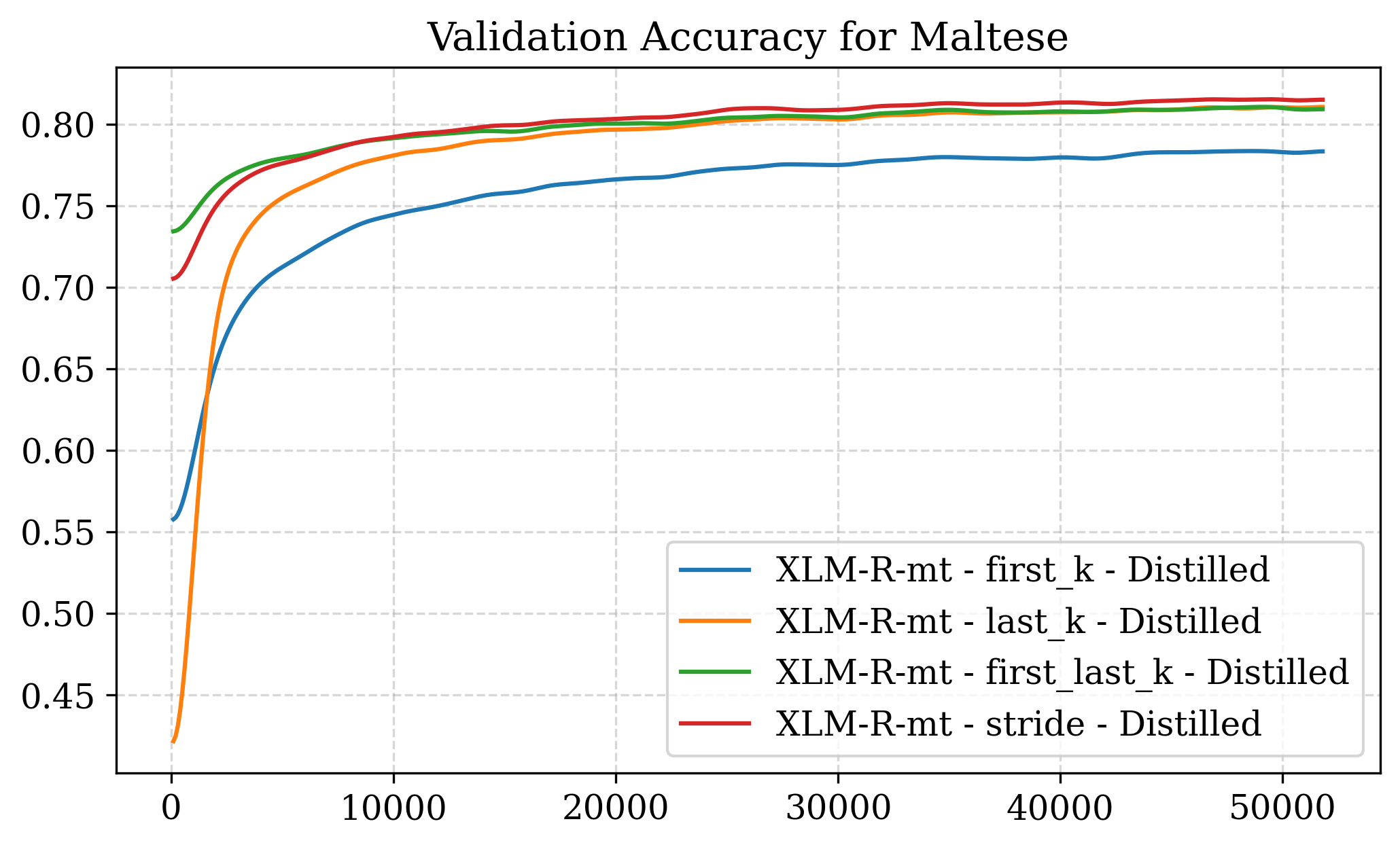}
        \caption{Maltese}
        \label{fig:init_kd_mt_xlm}
    \end{subfigure}
    \hfill
    \begin{subfigure}{0.48\linewidth}
        \centering
        \includegraphics[width=\linewidth]{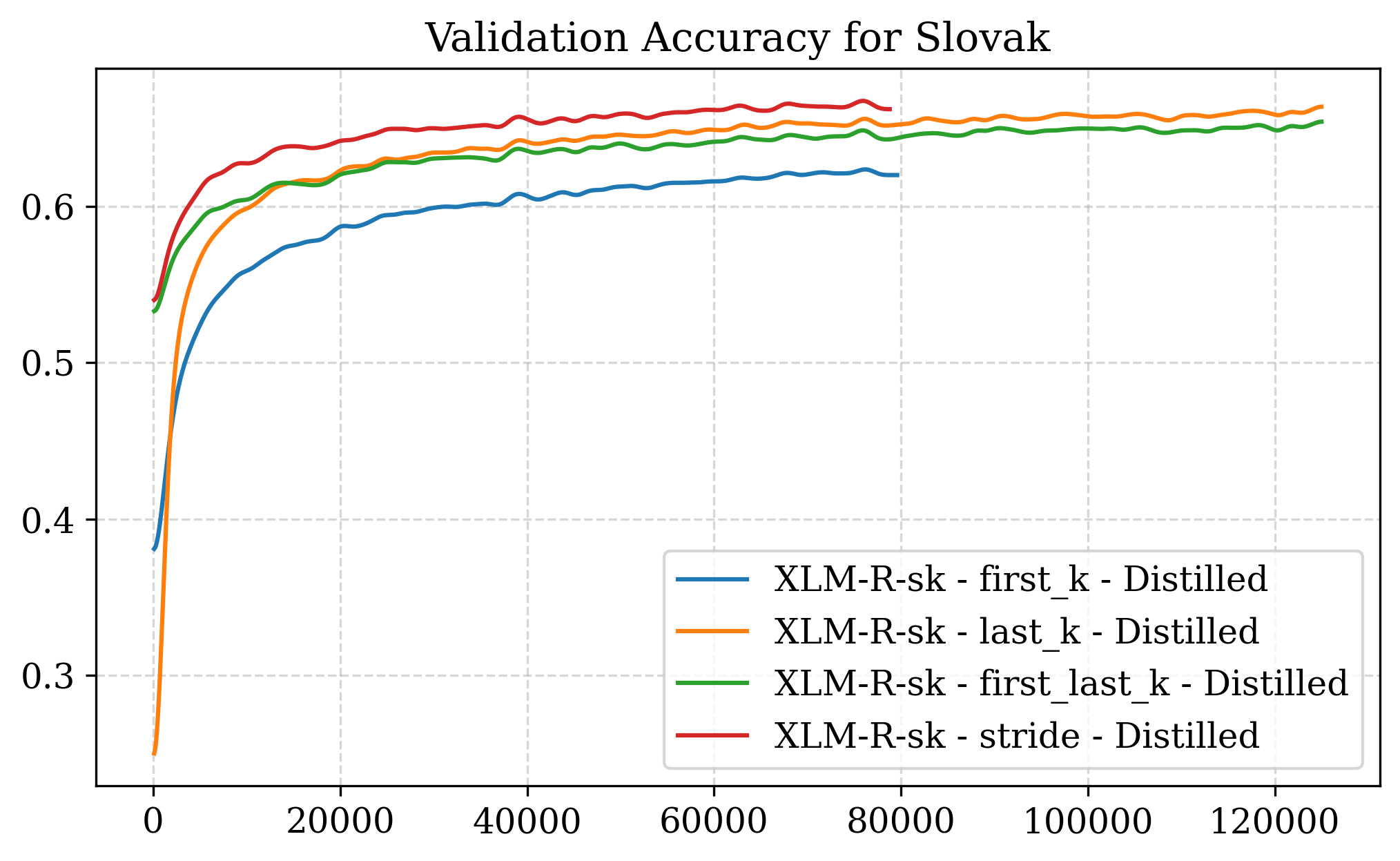}
        \caption{Slovak}
        \label{fig:init_kd_sk_xlm}
    \end{subfigure}
    
    \vspace{0.5cm} 

    \begin{subfigure}{0.48\linewidth}
        \centering
        \includegraphics[width=\linewidth]{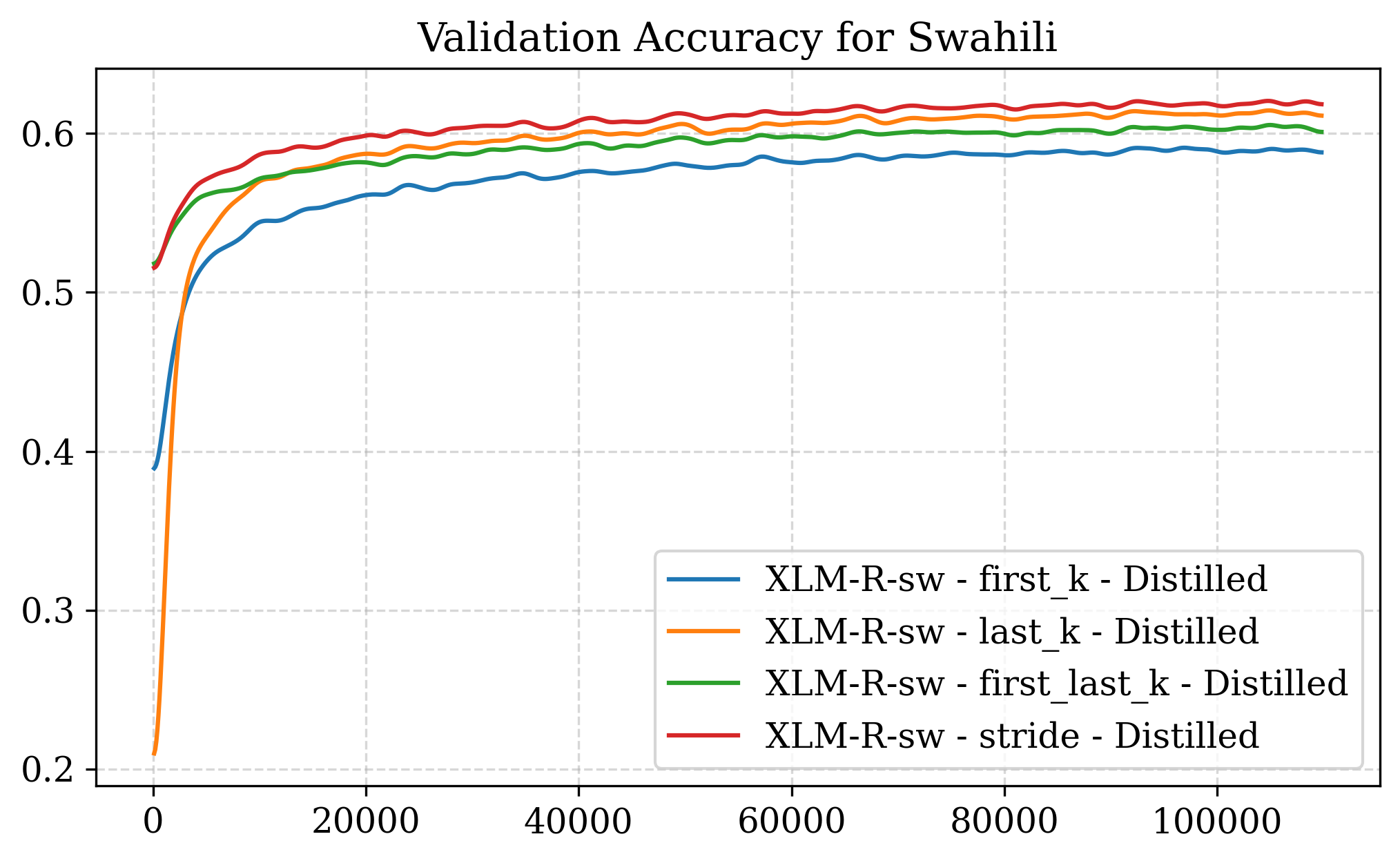}
        \caption{Swahili}
        \label{fig:init_kd_sw_xlm}
    \end{subfigure}

    \caption{Validation accuracy for various initialization strategies for XLM-R.}
    \label{fig:init_kd_xlm}
\end{figure}

\newpage
\section{SVD vs. Truncation for Hidden Size Reduction}
\label{app:svd}

\begin{figure}[h]
    \centering
    \centering
    \includegraphics[width=\linewidth]{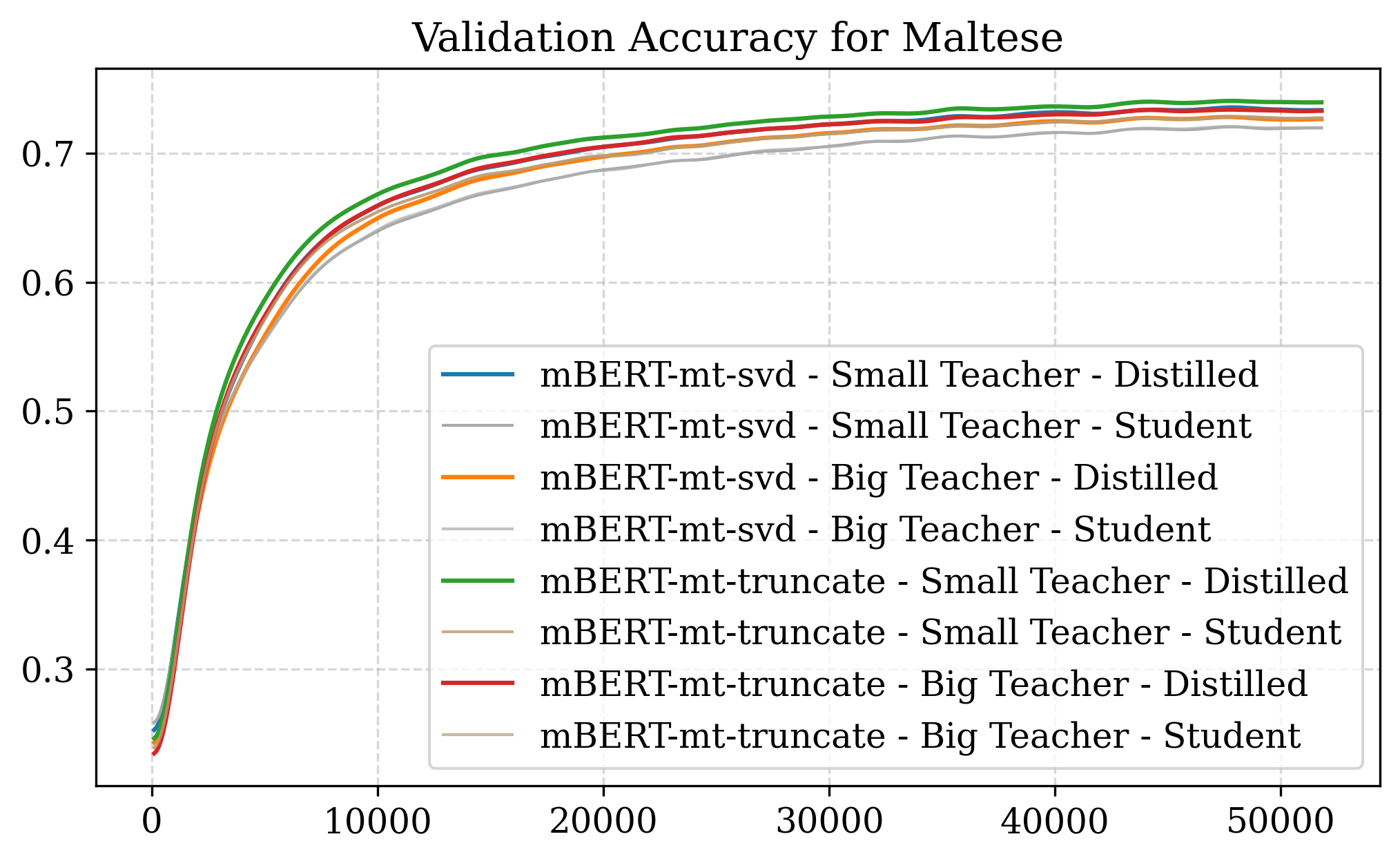}
    \caption{Validation accuracy comparing SVD vs. first-$k$ truncation for hidden size reduction to 312. ``Small teacher'' refers to the layer-compressed (6-layer) model; ``Big teacher'' is the original 12-layer language-adapted model. Truncation consistently outperforms SVD regardless of teacher size.}
    \label{fig:svd_trunc}
\end{figure}

\section{Alpha Parameter in Knowledge Distillation}
\begin{figure}[h]
    \centering
    \begin{subfigure}{0.48\linewidth}
        \centering
        \includegraphics[width=\linewidth]{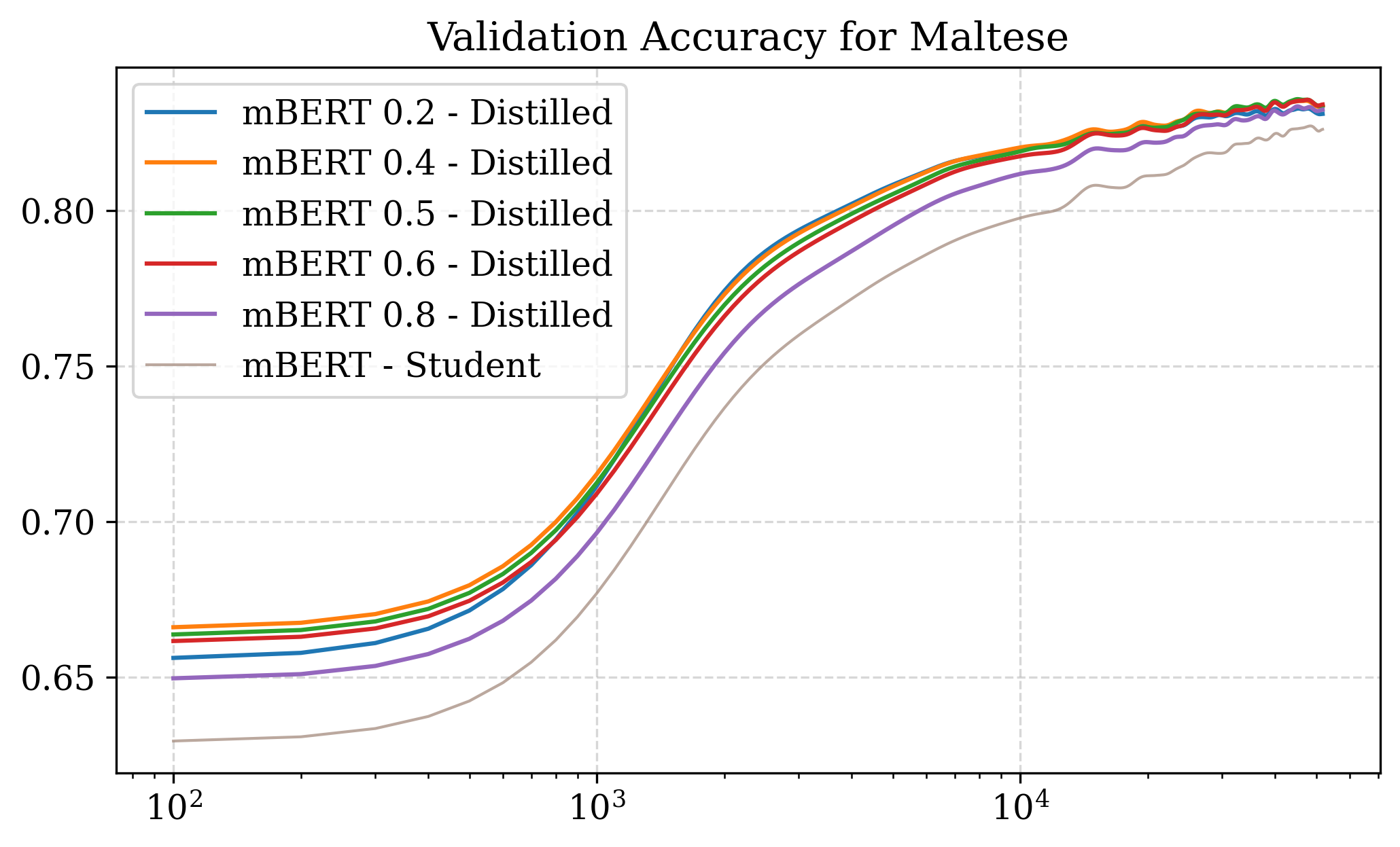}
        \caption{mBERT}
        \label{fig:alpha_mbert}
    \end{subfigure}
    \hfill
    \begin{subfigure}{0.48\linewidth}
        \centering
        \includegraphics[width=\linewidth]{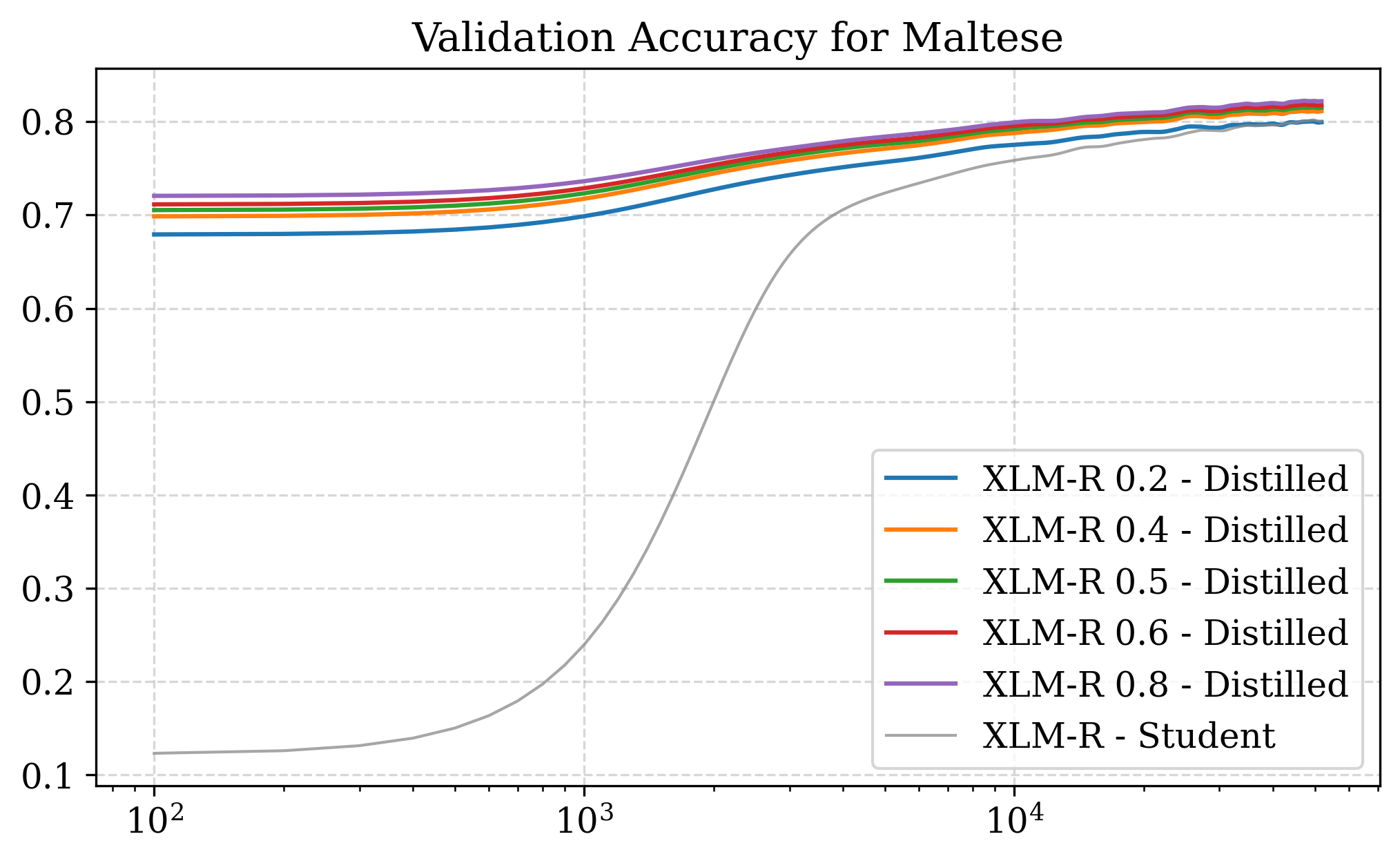}
        \caption{XLM-R}
        \label{fig:alpha_xlm}
    \end{subfigure}

    \caption{Validation accuracy curves showing the impact of the alpha parameter on knowledge distillation performance for mBERT and XLM-R on Maltese with the last $k$ and stride initialization strategies for the two models respectively.}
    \label{fig:alpha}
\end{figure}

We find that the $\alpha$ parameter does not have a significant impact on mBERT during pre-training, with $\alpha = 0.5$ yielding consistently good results. For XLM-R, higher values of $\alpha$ (i.e., 0.6 and 0.8), which reduce the strength of the distillation effect, show slightly improved validation accuracy trends compared to lower values. In our experiments, we adopt the default setting of $\alpha = 0.5$, leaving a more comprehensive exploration of optimal values across different languages, dataset sizes, and model architectures to future work.

\newpage
\section{Vocabulary Reduction Analysis}
\label{app:vocab}
\begin{figure}[h]
    \centering
    \begin{subfigure}{0.48\linewidth}
        \centering
        \includegraphics[width=\linewidth]{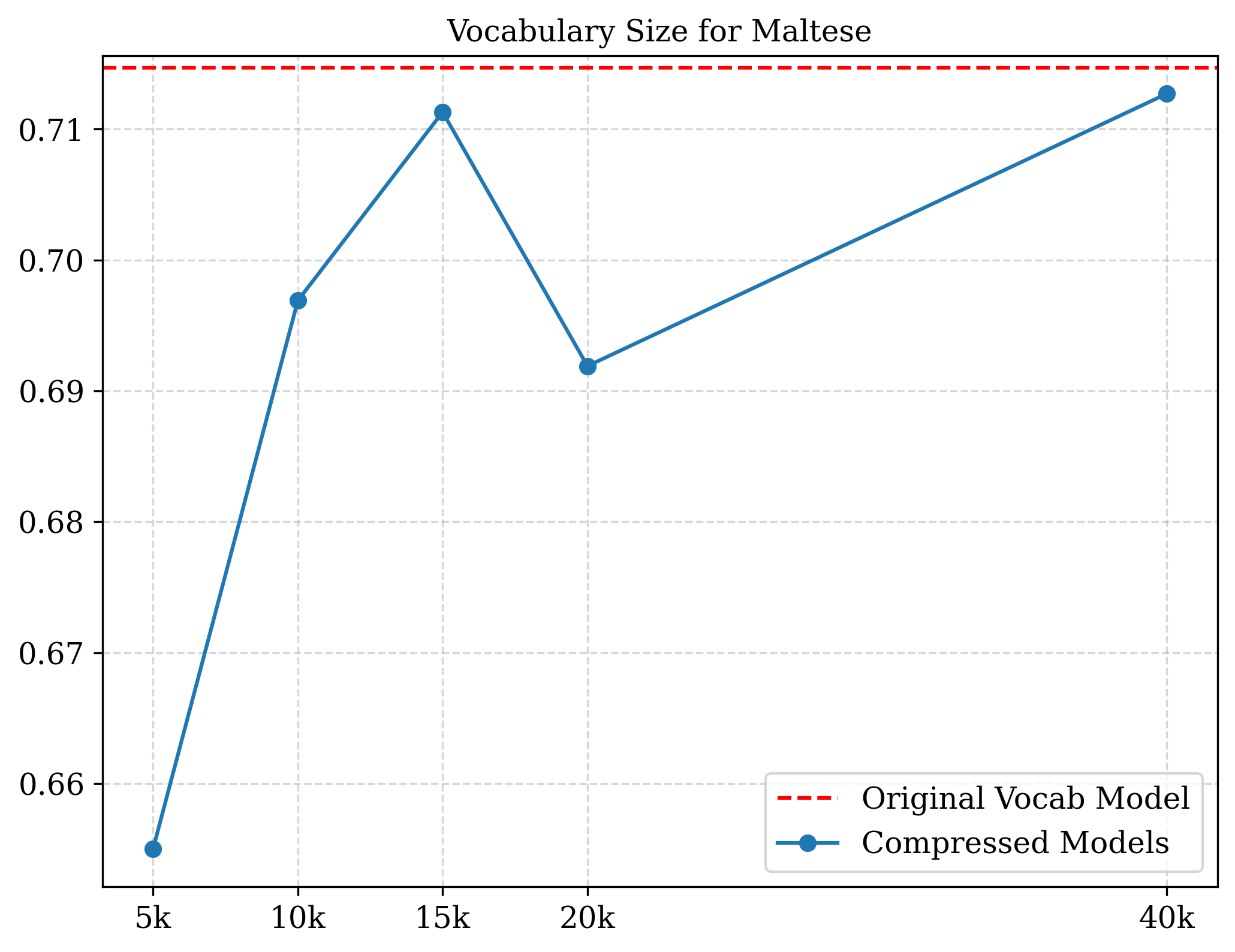}
        \caption{Maltese}
        \label{fig:vocab_mt}
    \end{subfigure}
    \hfill
    \begin{subfigure}{0.48\linewidth}
        \centering
        \includegraphics[width=\linewidth]{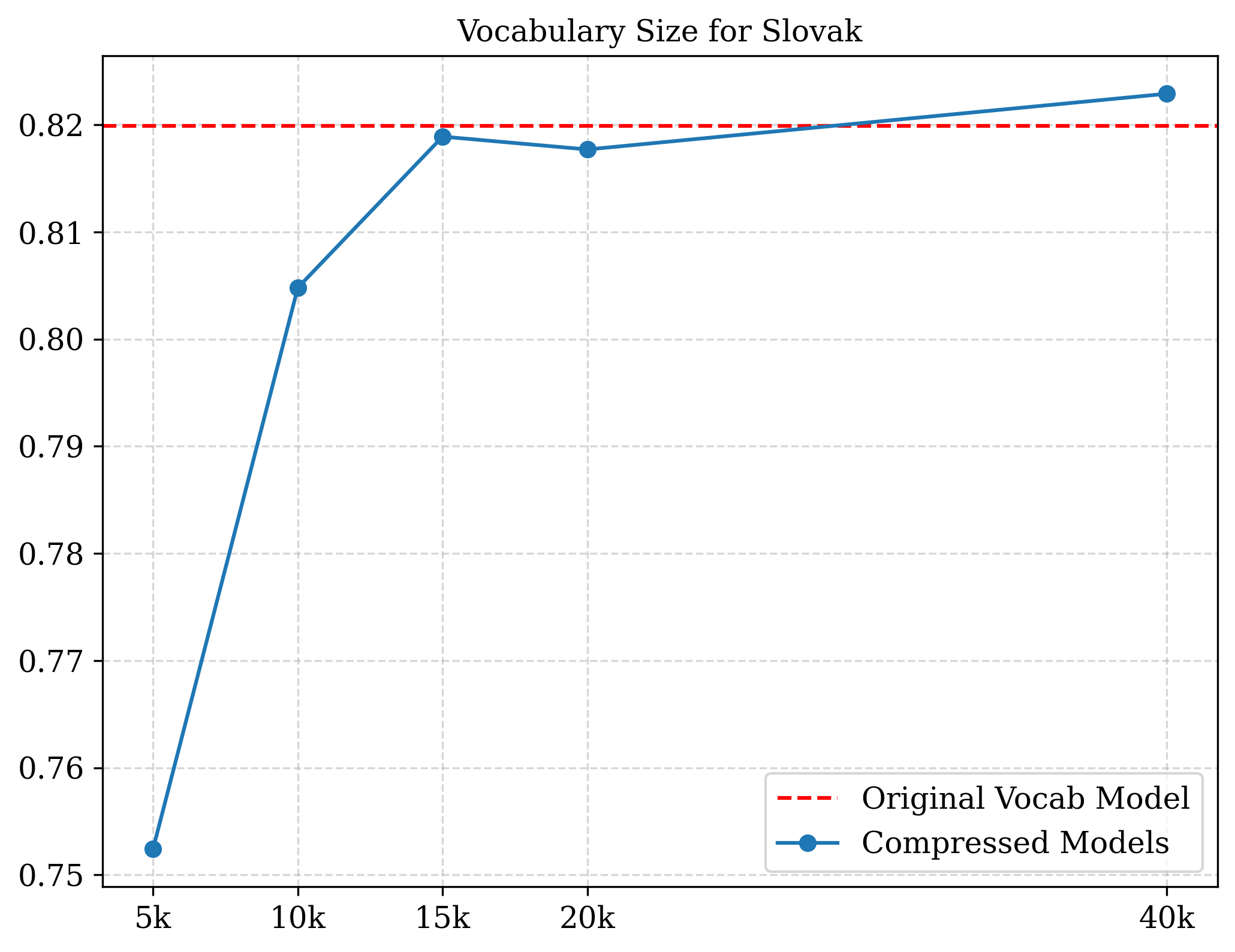}
        \caption{Slovak}
        \label{fig:vocab_sk}
    \end{subfigure}
    
    \vspace{0.5cm} 

    \begin{subfigure}{0.48\linewidth}
        \centering
        \includegraphics[width=\linewidth]{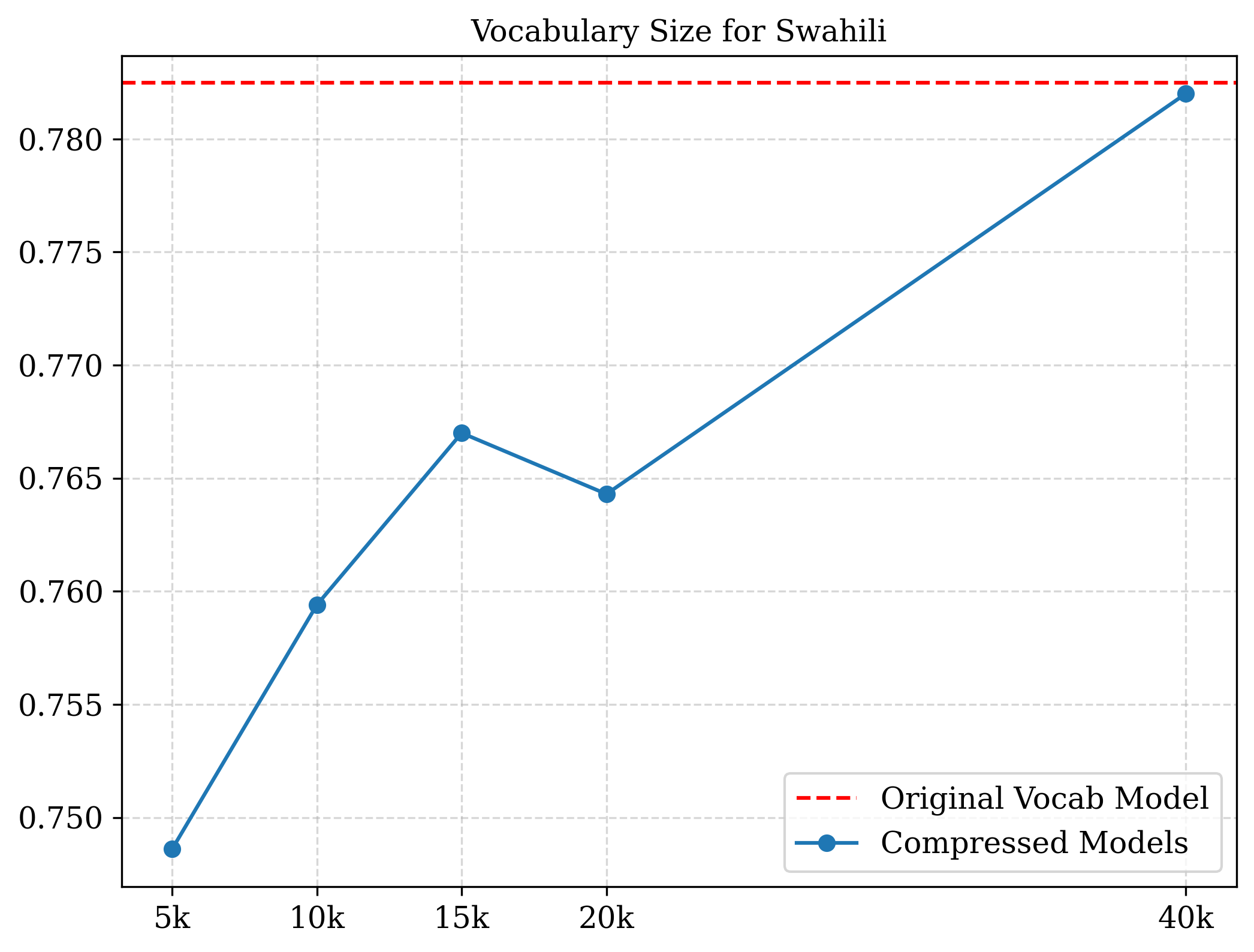}
        \caption{Swahili}
        \label{fig:vocab_sw}
    \end{subfigure}

    \caption{Impact of vocabulary reduction on TC performance for mBERT models reduced to a hidden size of 312.}
    \label{fig:vocab}
\end{figure}

\section{Knowledge Distillation Data Sizes}
\label{app:data}

\begin{table}[h]
\centering
\small
\setlength{\tabcolsep}{6pt}
\begin{tabular}{@{}l r r@{}}
\toprule
\textbf{Language} & \textbf{KD Data Size (MB)} & \textbf{FT Data Size (MB)} \\
\midrule
Maltese (mt)  & 238  & 188 \\
Slovak (sk)   & 535  & 1032 \\
Swahili (sw)  & 402  & 332 \\
\bottomrule
\end{tabular}
\caption{Dataset sizes for knowledge distillation (KD) and monolingual fine-tuning (FT) for each language. The language-adapted models are sourced from \citet{gurgurov2025smallmodelsbigimpact}, and the FT data sizes are as reported by them.}
\label{tab:kd_ft_data_sizes}
\end{table}


\newpage
\section{Downstream Task Data Sizes}

\begin{table}[h]
\centering
\small
\begin{tabular}{@{}l ccc@{}}
\toprule
\textbf{Language} & Train & Validation & Test \\
\midrule
\multicolumn{4}{c}{\textbf{Text Classification (TC)}} \\
\midrule
Maltese (mt) & 701 & 99 & 204 \\
Slovak (sk) & 701 & 99 & 204 \\
Swahili (sw) & 701 & 99 & 204 \\
\midrule
\multicolumn{4}{c}{\textbf{Sentiment Analysis (SA)}} \\
\midrule
Maltese (mt) & 595 & 85 & 171 \\
Slovak (sk) & 3560 & 522 & 1042 \\
Swahili (sw) & 738 & 185 & 304 \\
\midrule
\multicolumn{4}{c}{\textbf{Named Entity Recognition (NER)}} \\
\midrule
Maltese (mt) & 100 & 100 & 100 \\
Slovak (sk) & 20000 & 10000 & 10000 \\
Swahili (sw) & 1000 & 1000 & 1000 \\
\midrule
\multicolumn{4}{c}{\textbf{Part of Speech Tagging (POS)}} \\
\midrule
Maltese (mt) & 1123 & 433 & 518 \\
Slovak (sk) & 8483 & 1060 & 1061 \\
Swahili (sw) & 675 & 134 & 539 \\
\bottomrule
\end{tabular}
\caption{Fine-tuning data sizes for each task (Text Classification, Sentiment Analysis, Named Entity Recognition, Part of Speech Tagging) showing train, validation, and test splits across Maltese, Slovak, and Swahili.}
\label{tab:fine_tuning}
\end{table}

\section{Downstream Task Hyperparameters}
\label{app:hyper}
\begin{table}[h]
\centering
\small
\begin{tabular}{@{}l cccc@{}}
\toprule
\textbf{Hyperparameter} & \textbf{TC} & \textbf{SA} & \textbf{NER} & \textbf{POS} \\
\midrule
Learning rate & 1e-4 & 1e-4 & 3e-4 & 3e-4 \\
Batch size & 16 & 16 & 64 & 64 \\
Epochs & 20 & 20 & 100 & 100 \\
Maximum length & 256 & 256 & 512 & 512 \\
\bottomrule
\end{tabular}
\caption{Hyperparameters for task adapter fine-tuning across Text Classification (TC), Sentiment Analysis (SA), and Named Entity Recognition (NER) tasks.}
\label{tab:hyperparameters}
\end{table}



\section{Adapter Trainable Parameter Counts}
\label{app:red_factor}
\begin{table}[h]
\centering
\small
\setlength{\tabcolsep}{5pt}
\begin{tabular}{@{}lrr@{}}
\toprule
\multirow{2}{*}{\textbf{Model Configuration}} & \multicolumn{2}{c}{\textbf{Task Adapter Size}} \\
\cmidrule(l){2-3}
& \textbf{mBERT} & \textbf{XLM-R} \\
\midrule
Base & 894,528 & 894,528 \\
Base-[mt, sk, sw] & 894,528 & 894,528 \\
* KD layer red. ×2 & 447,264 & 447,264 \\
* inter. layer red. → 2048 & 447,264 & 447,264 \\
\midrule \midrule
* KD hid. size red. → 564 & 240,474 & 240,474 \\
* vocab. red. → 40k & 240,474 & 240,474 \\
\midrule \midrule
* KD hid. size red. → 456 & 156,120 & 156,120 \\
* vocab. red. → 40k & 156,120 & 156,120 \\
\midrule \midrule
* KD hid. size red. → 312 & 73,122 & 73,122 \\
* vocab. red. → 40k & 73,122 & 73,122 \\
\bottomrule
\end{tabular}
\caption{Task adapter parameter sizes across different model compression configurations for mBERT and XLM-R \textbf{with the default reduction factor of 16}. When the hidden size is reduced, adapter input/output dimensions decrease proportionally. When the layer count is reduced, fewer adapters are added to the model. All other parameters use the default settings for the Sequential Bottleneck adapter as implemented in AdapterHub.}
\label{tab:adapter-size}
\end{table}

\begin{figure}[h]
    \centering
    \begin{subfigure}{\linewidth}
        \centering
        \includegraphics[width=0.8\linewidth]{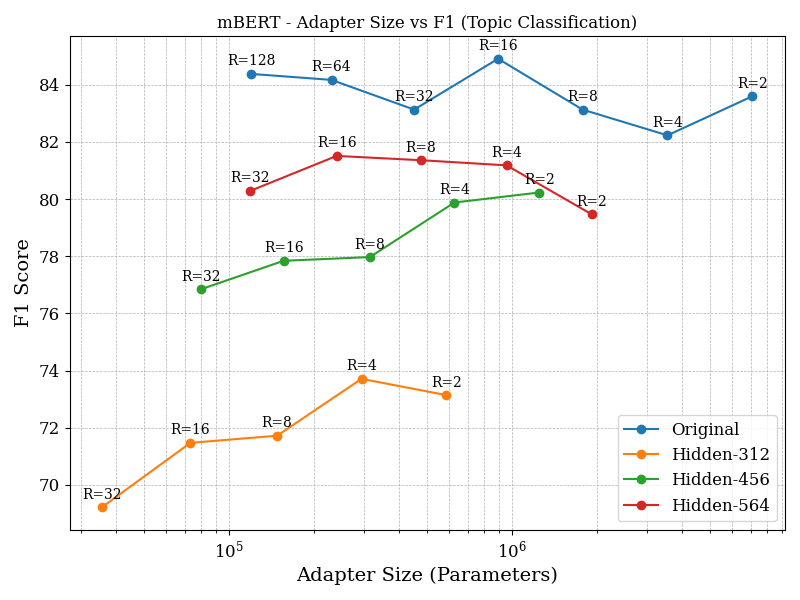}
        \caption{mBERT}
        \label{fig:r_mbert}
    \end{subfigure}
    
    \vspace{0.5em}

    \begin{subfigure}{\linewidth}
        \centering
        \includegraphics[width=0.8\linewidth]{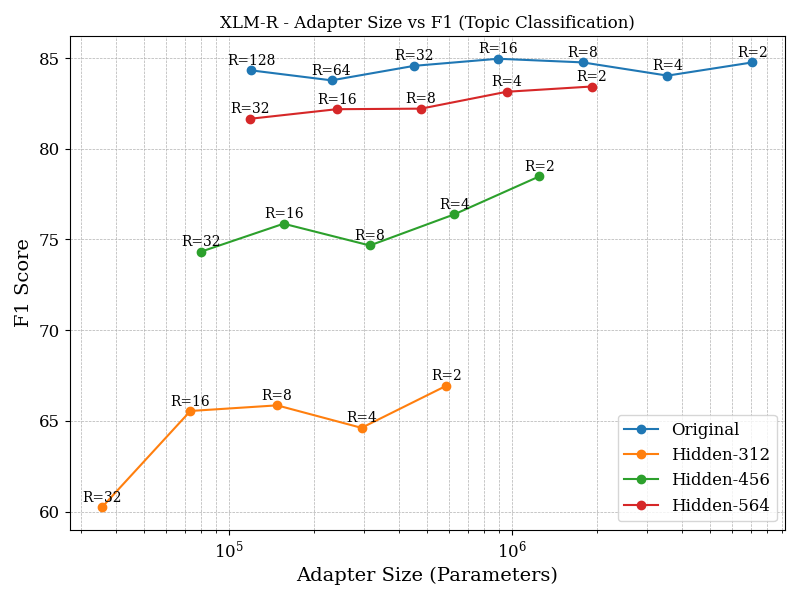}
        \caption{XLM-R}
        \label{fig:r_xlm}
    \end{subfigure}

    \caption{Performance of models on TC for Maltese with varying adapter capacity for mBERT and XLM-R.}
    \label{fig:red}
\end{figure}

To examine whether the constrained task adapter capacity, as shown in Table~\ref{tab:adapter-size}, impacts downstream performance in compressed models, we vary the reduction factor $r$, thereby increasing adapter size (see Figure~\ref{fig:red}). We train task adapters on top of both full adapted models and hidden-size reduced models (564, 456, and 312). For the smallest models (456 and 312), we observe that increasing adapter capacity ($r$=2) leads to improved performance. However, this increase is unnecessary for larger mBERT variants (full and 564), while still beneficial for all small XLM-R models. These results suggest that for smaller models, increasing adapter capacity can yield modest performance gains. Tables~\ref{tab:performance} and~\ref{tab:performance-mbert} report results using the default reduction rate of 16.

\newpage

\begin{table*}[t]
\section{Downstream Results for mBERT}
\label{app:mbert}
\centering
\small
\resizebox{\textwidth}{!}{
\begin{tabular}{@{}l r r  @{\hspace{0.6em}}rrrr rrrr rrrr r@{}} 
\toprule
\multirow{2}{*}{\textbf{Compression Stage}} & \multirow{2}{*}{\textbf{Params}} & \multirow{2}{*}{\textbf{Size}} & \multicolumn{12}{c}{\textbf{Task Performance (F1)}} & \multirow{2}{*}{\textbf{Avg}} \\
\cmidrule(lr){4-15}
& & & \multicolumn{4}{c}{\textbf{Maltese}} & \multicolumn{4}{c}{\textbf{Slovak}} & \multicolumn{4}{c}{\textbf{Swahili}} \\ 
\cmidrule(lr){4-7} \cmidrule(lr){8-11} \cmidrule(lr){12-15}
& & & TC & SA & NER & POS & TC & SA & NER & POS & TC & SA & NER & POS \\ 
\midrule
\multicolumn{15}{l}{\textit{Baselines}} \\
\quad Multilingual & 179M & 179M & 68.7 & 65.8 & 60.0 & 89.0 & 85.3 & 92.0 & 91.4 & 97.0 & 69.6 & 64.6 & 83.8 & 87.6 & 79.6 \\ 
\rowcolor{gray!10}
\quad Language-adapted & 179M & 179M & 84.9 & 73.6 & 65.0 & 94.0 & 86.3 & 91.9 & 90.4 & 96.9 & 86.7 & 81.3 & 82.5 & 88.7 & \textbf{85.2} \\ 
\midrule
\multicolumn{15}{l}{\textit{Compression Pipeline (minimal degradation)}} \\
\quad Layer reduction & 135M (-25\%) & 135M & 80.1 & 73.9 & 59.0 & 93.2 & 85.4 & 90.4 & 87.4 & 96.9 & 82.8 & 77.3 & 80.7 & 88.4 & 83.0 \\ 
\quad + FFN pruning & 126M (-30\%) & 126M & 79.0 & 74.7 & 58.1 & 92.7 & 85.3 & 90.2 & 88.5 & 96.7 & 83.2 & 75.9 & 79.8 & 88.5 & 82.7 \\ 
\quad + Hidden 564 & 90M (-50\%) & 90M & 79.5 & 70.2 & 61.1 & 92.6 & 83.4 & 90.5 & 88.1 & 96.3 & 83.5 & 76.1 & 79.7 & 88.4 & 82.5 \\ 
\rowcolor{green!20}
\quad + Vocabulary & 45M (-75\%) & 45M & 80.2 & 70.8 & 61.1 & 92.5 & 83.5 & 90.7 & 87.7 & 96.3 & 84.3 & 76.0 & 80.3 & 88.6 & \textbf{82.7} \\ 
\midrule
\multicolumn{15}{l}{\textit{Further compression (moderate degradation)}} \\
\quad + Hidden 456 & 71M (-60\%) & 71M & 80.2 & 70.1 & 57.2 & 92.1 & 83.9 & 90.4 & 87.5 & 95.9 & 85.1 & 78.6 & 80.3 & 88.3 & 82.5 \\ 
\quad + Vocabulary & 35M (-80\%) & 35M & 81.0 & 69.7 & 55.9 & 92.0 & 84.2 & 90.4 & 87.4 & 96.0 & 83.0 & 78.5 & 79.8 & 88.4 & 82.2 \\ 
\midrule
\multicolumn{15}{l}{\textit{Maximum compression (higher degradation)}} \\
\rowcolor{orange!15}
\quad + Hidden 312 & 48M (-73\%) & 48M & 73.1 & 72.0 & 39.5 & 90.3 & 80.9 & 90.4 & 86.5 & 95.5 & 81.8 & 76.5 & 79.6 & 87.7 & 79.5 \\ 
\rowcolor{orange!15}
\quad + Vocabulary & 23M (-87\%) & 23M & 73.0 & 72.1 & 40.4 & 90.2 & 81.9 & 90.1 & 86.2 & 95.3 & 81.7 & 76.0 & 77.1 & 87.7 & 79.3 \\ 
\bottomrule
\end{tabular}
}
\caption{Progressive compression of mBERT. Stages are grouped by degradation level. Highlighted rows indicate the baseline (gray) and optimal compression point (green, 75\% reduction with 2.5\% drop). Maximum compression rows (red) show significant degradation (5.9\% drop). TC=Topic Classification, SA=Sentiment Analysis, NER=Named Entity Recognition, POS=Part-of-Speech Tagging. F1 scores averaged over 3 runs.}
\label{tab:performance-mbert}
\end{table*}

\end{document}